\documentclass[conference]{IEEEtran}
\IEEEoverridecommandlockouts
\usepackage{cite}
\usepackage{amsmath,amssymb,amsfonts}
\usepackage{algorithmic}
\usepackage{graphicx}
\usepackage{textcomp}
\usepackage{xcolor}
\def\BibTeX{{\rm B\kern-.05em{\sc i\kern-.025em b}\kern-.08em
    T\kern-.1667em\lower.7ex\hbox{E}\kern-.125emX}}

\usepackage[switch]{lineno}
\usepackage{multirow}
\usepackage{verbatim}
\usepackage{array}
\usepackage{pgfplots}
\usepackage{algorithm}
\usepackage{algorithmic}
\usepackage{listings}
\usepackage{booktabs}
\usepackage{hyperref}

\newcolumntype{?}{!{\vrule width 2pt}}
\makeatletter
\newcommand{\thickhline}{%
    \noalign {\ifnum 0=`}\fi \hrule height 2pt
    \futurelet \reserved@a \@xhline
}
\newcolumntype{"}{@{\hskip\tabcolsep\vrule width 1pt\hskip\tabcolsep}}

\usepackage{xcolor}

\usepackage{xcolor}

\newif\ifrevision
\revisiontrue   % set \revisionfalse to disable

\definecolor{revfg}{RGB}{0,102,204} % blue (change as you like)

\begin{document}

\title{ZTab: Domain-based Zero-shot Annotation for Table Columns%
\thanks{Published in the Proceedings of the IEEE 42nd International Conference on Data Engineering (ICDE 2026)}
}

% \title{ZTab: Domain-based Zero-shot Annotation for Table Columns\\
% %\thanks{Identify applicable funding agency here. If none, delete this.}
% }

\author{\IEEEauthorblockN{Ehsan Hoseinzade}
\IEEEauthorblockA{\textit{School of Computing Science} \\
\textit{Simon Fraser University}\\
Burnaby, Canada \\
ehoseinz@sfu.ca \\
%ORCID: 0000-0002-6742-7135
}
\and
\IEEEauthorblockN{Ke Wang}
\IEEEauthorblockA{\textit{School of Computing Science} \\
\textit{Simon Fraser University}\\
Burnaby, Canada \\
wangk@cs.sfu.ca \\
%ORCID: 0000-0002-8021-4951
}
%\and
%\and
%\and
% \and
}

\maketitle

\begin{abstract}
This study addresses the challenge of automatically detecting semantic column types in relational tables, a key task in many real-world applications. Zero-shot modeling eliminates the need for user-provided labeled training data, making it ideal for scenarios where data collection is costly or restricted due to issues such as privacy concerns. However, existing zero-shot models suffer from poor performance in the case of a large number of semantic column types or classes, poor understanding of tabular structures, and privacy risks arising from dependency on high-performance closed-source LLMs. We introduce ZTab, a domain-based zero-shot framework, to address both performance and zero-shot requirements. ZTab considers a domain configuration given by a set of predefined semantic types, plus sample table schemas based on such types, fine-tunes an annotation LLM using pseudo-tables generated for sample table schemas. ZTab is domain-based zero-shot in that it does not depend on user-specific labeled training data; therefore, no retraining is needed for a test table coming from a similar domain. We describe three cases for domain-based zero-shot. The domain configuration of ZTab provides a trade-off between the extent of zero-shot and the annotation performance: for a ``universal domain" that contains all semantic types, domain-based zero-shot will approach ``pure" zero-shot; on the other hand, a ``specialized domain" that contains semantic types for a specific application will enable better zero-shot performance within that domain. The source code and datasets are available at \href{https://github.com/hoseinzadeehsan/ZTab}{https://github.com/hoseinzadeehsan/ZTab}. 

\end{abstract}


\section{Introduction}
In real-world scenarios, column headers of tables are often missing, or generic (e.g., ``Value"), or auto-generated (e.g., ``col1"), especially in user-generated spreadsheets, web tables, and automated ETL pipelines \cite{kang2003schema,trabelsi2020semantic,singha2023tabular}.
Column type annotation aims to automatically identify or tag the semantic types or classes of the columns for such tables\footnote{Semantic types such as Person Name are different from atomic types such as String. Identifying atomic types given values is trivial, but identifying semantic types given values is challenging}, which is crucial for different information retrieval tasks like data integration \cite{hai2023data}, data cleaning \cite{limaye2010annotating,kandel2011wrangler}, schema matching \cite{rahm2001survey}, and data discovery \cite{fernandez2018seeping,fernandez2018aurum}. In particular, table discovery in data lakes often relies on semantic signals such as column types to support search and integration \cite{fan2023table}.
 One emerging application is automatically tagging sensitive columns in a table, such as personal information, before deciding what information can be released. 
Supervised learning-based methods \cite{hul2019sherlock, zhang2019sato, deng2022turl, suhara2022annotating, sun2023reca, hoseinzade2024graph} have shown promising results when the test data belong to the same domain and distribution as the training data, a setting referred to as In-Domain Generalization. These models primarily leverage BERT’s pre-training on large-scale textual corpora, fine-tuning it for labeled tabular training data by adding task-specific output layers to classify each column into a predefined set of semantic types. However, these models face significant limitations due to their dependency on user-provided labeled tabular training data in the following scenarios:

\textit{Data Availability:} Collecting high-quality labeled tabular training data is a resource-intensive and time-consuming process. In many cases, such data might not exist in the required format, or it could be confidential since privacy concerns and regulations such as HIPAA and GDPR impose restrictions on sharing sensitive data, making it nearly impossible to collect and use training data in such applications. 

\textit{Cross-Domain Generalization:} Supervised models often suffer from domain bias, i.e.,  perform poorly on datasets from different but related domains. For example, models like HNN \cite{chen2019learning} and ColNet \cite{chen2019colnet} trained on the T2D \cite{t2d} from Webtables show weak performance on Limaye \cite{limaye2010annotating} and Efthymiou \cite{efthymiou2017matching} from Wikipedia, despite shared classes \cite{chen2019learning}. This issue, known as domain shift, arises from differences in data distributions.

\textit{Cross-Ontology Generalization:} These models are inherently designed for a fixed set of class labels, limiting their ability to handle scenarios involving labels from a different ontology. For example, a model trained on datasets annotated with Schema.org’s ``Person'' type would fail to classify columns annotated with DBpedia’s ``Human'' type in a healthcare integration task. This mismatch arises because the model cannot adapt to new ontologies without retraining, making it ineffective for cross-ontology column type annotation.

Large Language Models (LLMs) have been proposed as a solution to these challenges through their ability to perform zero-shot column type annotation without the need for specific, labeled tabular datasets \cite{kayali2024chorus,korini2023column,feuer2024archetype,zhang2024jellyfish,zhang2024tablellama, xiao2025cents} (Figure \ref{fig: overall ZTab}(a)). 
% These models, pre-trained on extensive textual corpora, can generalize across various domains, making them attractive for scenarios where labeled data is difficult to obtain.  
However, despite their potential, current LLM-based zero-shot models for column type annotation suffer from the following limitations:

\begin{figure*}[t]
\centering
\includegraphics[width=1\textwidth]{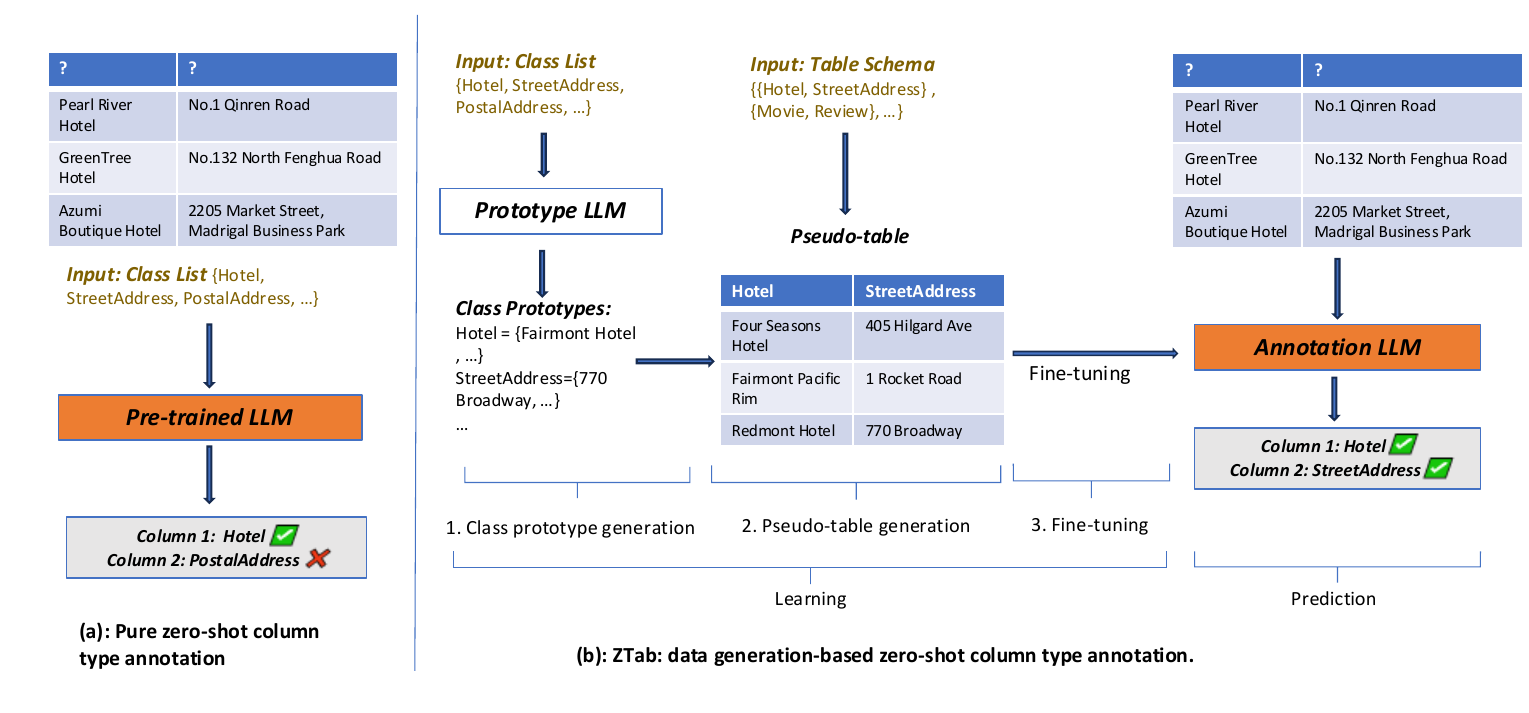} 
%\vspace{-8mm}
\caption{Comparison between (a) pure zero-shot column type annotation and (b) domain-based zero-shot ZTab. ZTab takes a class list and table schema collection as inputs, generates class prototypes (1) and pseudo-tables based on class prototypes (2), and fine-tunes an annotation LLM (3).}
\label{fig: overall ZTab}
\vspace{-5mm}
\end{figure*}

\textbf{Performance:} The performance of LLMs in zero-shot classification often falls short, particularly when dealing with closely related classes like ``addressRegion," ``addressLocality," ``streetAddress," and ``PostalAddress" \cite{feuer2024archetype} as they struggle to learn the fine differences between them. The issue becomes worse as the number of classes increases \cite{vandemoortele2025haystack}. The other difficult case is semantic types that are numeric, such as rank and position \cite{hul2019sherlock}, where numeric values provide little information about their semantic types.  

\textbf{Structure:} LLMs, being primarily pre-trained on unstructured textual data, struggle to learn the structural relationships between columns within tables \cite{li2024table,sui2024table}. Consequently, zero-shot models based on LLMs are less effective at capturing table-specific details like values in the same column or rows, unlike supervised models, which directly learn them from tabular training data. 

\textbf{Privacy:} current LLM-based zero-shot models depend on
powerful closed-source LLMs like GPT for good performance, posing privacy risks because such LLMs require sensitive table information to be sent to the third party owning the LLM at inference time.  

\textit{\textit{The question is how to balance between the zero-shot requirement and the performance requirement for table column annotation, and additionally, the data privacy requirement.}} We propose a novel \textbf{domain-based zero-shot framework}, \textbf{ZTab}, as a solution to this question. ZTab considers a ``domain" defined by a set of pre-defined semantic types or classes (e.g., Country, Capital, GDP), plus sample table schemas based on such classes. The class set provides the domain information about the application, and table schemas provide sample table structures, both at the schema level without involving actual data. ZTab uses a \textit{prototype LLM} to generate representative values for each class, called class prototypes (e.g., Canada, UK, France for the class Country), and fine-tunes an \textit{annotation LLM} for column annotation using pseudo-tables generated from table schemas and class prototypes. Supervised fine-tuning addresses the above Performance and Structure issues. Privacy issue is addressed since no user-specific training data beyond table schemas and classes 
is needed.

We consider two variants of ZTab. \textbf{ZTab-privacy}, used when user privacy is a concern, denotes ZTab that uses open-source prototype LLM and annotation LLM; therefore, fine-tuning and inference can be performed locally, exposing no real user data to a third party. 
\textbf{ZTab-performance}, used when performance outweighs privacy, denotes ZTab that 
utilizes more powerful closed-source LLMs (like GPT models). Figure \ref{fig: overall ZTab}(b) shows how ZTab-performance works, where pseudo-tables are generated before fine-tuning a closed-source annotation LLM. For ZTab-privacy, pseudo-tables are generated adaptively at each epoch of fine-tuning for more diversity of data. See more details in Section \ref{sec:ztab}.

The key contributions of ZTab are as follows:
\begin{itemize}
    \item  \textbf{Domain-based zero-shot framework}: ZTab is domain-based zero-shot in that no retraining (i.e., fine-tuning) is needed for test data coming from a similar domain. This includes three important cases: (1) \textbf{In-Domain} generalization, where the test table comes from the same class list (but not necessarily the same data distribution) on which ZTab is trained, (2) \textbf{Cross-Domain} generalization where the test table comes from a more restricted class list than that on which ZTab is trained, and (3) \textbf{Cross-Ontology} generalization where the test table comes from the same class list but the classes are derived from a different ontology. Meanwhile, fine-tuning step helps addressing the performance issue of pure zero-shot learning.

    \item \textbf{Robust zero-shot performance:}  For the In-Domain setting, ZTab-performance achieves the highest results, outperforming the strongest GPT-4o baseline by $4.5\%$. ZTab-privacy, using open-source models with 7–8 billion parameters, improves over the strongest open-source baseline by $23.5\%$ and is comparable to GPT-4o baselines while fully preserving data privacy. For Cross-Domain and Cross-Ontology settings,   
    ZTab-performance improves existing zero-shot baselines by at least $2.7\%$ and $3.8\%$, respectively, and ZTab-privacy improves over the strongest open-source baselines by $1.4\%$ and $9.5\%$, respectively, and shows performance comparable to the strongest GPT-4o-based model while fully preserving privacy. These results are reported in Table \ref{tab:model_comparison}.

    \item \textbf{A trade-off between the extent of zero-shot and annotation performance}: for a ``universal domain” that contains all semantic types, domain-based zero-shot will approach ``pure” zero-shot where no retraining is needed because every test table comes from a more restricted class list (i.e., the Cross-Domain setting); on the other hand, a ``specialized domain” that contains the semantic types for a specific scenario will enable better annotation within that domain. The user can obtain a desired trade-off between the extent of zero-shot and the annotation performance by choosing a proper domain configuration. 

\end{itemize}

\section{Related Works}
Recent column type annotation methods are typically grounded in two categories, i.e., supervised learning and zero-shot learning. The former requires labeled tabular training data, while the latter can perform annotation without this need for tabular training data.

\textbf{Supervised learning:} These methods heavily relies on tabular training data \cite{miao2023watchog}. They train/fine-tune either a \textit{deep learning} model or a \textit{language model}, usually BERT \cite{devlin2018bert}, for column type annotation. In the deep learning category, ColNet \cite{chen2019colnet} uses DBpedia cell value lookups to create examples and trains a CNN; HNN \cite{chen2019learning} models intra-column semantics, enhancing Colnet. Sherlock \cite{hul2019sherlock} employs statistical and textual features of a column using a neural network; SATO \cite{zhang2019sato} builds on Sherlock by modeling column dependencies using a CRF. In the language model category, TURL \cite{deng2020turl} is pre-trained for table understanding and then fine-tuned for column type annotation; Doduo \cite{suhara2022annotating} predicts all the columns of a table together by feeding the whole table to BERT; RECA \cite{sun2023reca} incorporates inter-table context information; 
GAIT \cite{hoseinzade2024graph} improves RECA by including intra-table information through a graph neural network. KGLink \cite{10597678} combines a knowledge graph with BERT to improve prediction.

\textbf{Zero-shot learning:} LLMs like GPT have been applied to column type annotation without the need for tabular training data \cite{kayali2024chorus,korini2023column,feuer2024archetype}. Korini et al. \cite{korini2023column} proposed converting tables into natural language and use ChatGPT for column annotation. Chorus \cite{kayali2024chorus} leverages GPT-3.5 for column type annotation by querying the entire table, but its applicability is limited to DBpedia properties. CENTS \cite{xiao2025cents} reduces inference cost by selecting representative table cells under a token budget while maintaining high accuracy across table understanding tasks. ArcheType \cite{feuer2024archetype} offers a broader framework for zero-shot annotation by using prompt serialization and label remapping, while also allowing fine-tuning of LLMs for better performance. Other models like Table-GPT \cite{li2024table}, TableLlama \cite{zhang2024tablellama}, and Jellyfish \cite{zhang2024jellyfish} fine-tune LLMs on table-related tasks to improve their alignment with tabular data. However, approaches using closed-source LLMs like GPT-3.5 raise privacy concerns due to data sharing with third-party GPT providers. 

In summary, supervised learning models are unsuitable when training data is not easily available due to privacy concerns, or when train and test sets come from different data distributions, or when syntactically different sets of labels are used, whereas zero-shot learning models' performance lags behind supervised models \cite{feuer2024archetype}. Our ZTab bridges this gap, i.e., eliminates the need for user-provided tabular training data while addressing the performance and privacy concerns.

\textbf{Data generation-based zero-shot learning:} \cite{li2023synthetic,ye2022zerogen,wang2021towards,tang2023does,gao2023self,zou2024fusegen} leverages LLMs to directly generate synthetic data for text classification, and TabGen \cite{berkovitch2025generating} prompts an LLM with a list of column headers to generate entire tables. These works do not consider column type annotation. 
Instead of prompting LLMs to generate entire tables, our ZTab prompts LLMs to generate only class prototypes (i.e., class-specific values) and builds pseudo-tables using table schemas and class prototypes adaptively in each epoch of fine-tuning. This approach is more efficient, easier to control in table size, and generates diverse tables on the same table schema. 

\section{Problem Statement}
We study the problem of \textit{column-based zero-shot column type annotation}. The task is to predict the semantic type of each column in a given table without column headers. In particular, we want to assign semantic types \( (c_1, c_2, \dots, c_n) \) to the columns of a given table \( T = (t_1, t_2, \dots, t_n) \) where each semantic type (or class) \( c_i \) is selected from a pre-defined set \( C_{pred} \). (While open-ended type prediction where the class set is not pre-defined is interesting, it is outside the scope of standard column type annotation and our work.) The solution to this problem has the learning phase that learns a model, and the deployment phase that applies the learned model to predict column semantic types for a given table. 

\textbf{Learning:} This phase has the following input:
% \begin{itemize}
(1) A set of semantic types \( C_{learn} = \{c_1, c_2, \dots, c_m\} \) to learn, such as \textit{Name}, \textit{Date}, or \textit{Condition}. (2) A set of table schemas \( S = \{S_1, S_2, \dots, S_k\} \), where each \( S_i = \{h_1, h_2, \dots, h_n\} \) represents the headers of the columns in a table and each \( h_i \) specifies the semantic type of a column and is from \( C_{\text{learn}} \). For example, \( S_i = \{\text{``Country", ``Locality"}\} \) represents a table having two columns of the semantic types ``Country" and ``Locality".  
% \end{itemize}
$C_{learn}$, called the \textit{learning domain}, describes all semantic types for an application and $S$ provides training samples for table structures, both at the schema level without involving actual column data. The learning phase aims to produce a model for predicting the semantic types of columns in a new header-less table that comes from a ``related domain", detailed below. 

\textbf{Deployment:}
At the deployment phase, the learned model is provided with a new table \( T \) containing \( n \) columns without headers and a candidate set of semantic types \( C_{pred} \), called the \textit{test domain}. The task is to annotate each column in \( T \) with a type from \( C_{pred} \). We consider three scenarios of the test domain \( C_{pred} \) where no retraining is needed (thus, zero-shot).
\begin{itemize}
    \item \textit{In-Domain Generalization}: In this case, \( C_{pred} = C_{learn} \), that is, the table \( T \) comes from the same domain as $C_{learn}$ on which the model is trained, 
    for example, both are from Wikipedia\footnote{https://www.wikipedia.org/}. This is the traditional scenario addressed by supervised learning. Note that the schema of $T$ is not required to be in $S$.

    \item \textit{Cross-Domain Generalization}: Unlike the previous scenario, the table \( T \) originates from a domain that is a subset of $C_{learn}$, i.e.,  $C_{pred} \subset C_{learn}$.    
    For example, $C_{learn}$ and \( S \) are from WebTables\footnote{https://webdatacommons.org/}, whereas \( T \) is from Wikipedia\footnotemark[1], which is a structured subset of WebTables.

    \item \textit{Cross-Ontology Generalization}: The table \( T \) comes from the same domain as  $C_{learn}$, but the semantic types in \( C_{pred} \) and \( C_{learn} \) are derived from different ontologies. For example, Schema.org\footnote{https://schema.org/} uses $C_{\text{learn}} = \{$ ``Person", ``Place", ``Organization"$\}$, whereas DBpedia\footnote{https://www.dbpedia.org/} uses $C_{\text{pred}} = \{$``Human", ``Location", ``Company"$\}$. This scenario requires the model to handle shifts in ontologies.

\end{itemize}

In the above three cases, this learning framework is domain-based zero-shot 
in the sense that no retraining is needed for any test table $T$. This is because the learning phase depends only on $C_{learn}$ and $S$, which are independent of actual training data. In fact, no retraining is needed even for a changed test domain $C_{pred}$, provided that it falls into one of these cases.

\begin{algorithm}[t]
\small
\caption{Learning Phase of ZTab-privacy}
\label{alg:learning}
\begin{algorithmic}[1]

\REQUIRE Set of classes $C_{learn}$, Table schema collection $S$, (open-source) Annotation LLM $M_a$, (open-source) Prototype LLM $M_p$, Schema sampling ratio $r$, Class prototype size $e$, row size $k$
\ENSURE Fine-tuned LLM $M_a$ \\
%\STATE \textcolor{blue}{\textit{class descriptions generation}}
\COMMENT{\textcolor{olive}{\textit{step 1: class prototypes generation}}}
\STATE $P$ $\leftarrow$ empty list
\FOR{each class $c_i$ in $C_{learn}$}
\STATE $p$ $\leftarrow$ \textbf{ClassPrototypeGeneration}($c_i$, $e$, $M_p$)
\STATE Add $p$ to $P$
\ENDFOR \\
\COMMENT{\textcolor{olive}{\textit{Step 2: handling missing classes}}}

\STATE $C_{missing} \leftarrow C_{learn} \setminus \{ \bigcup_{S_i \in S} S_i \}$

\STATE $S_{manual} \leftarrow \{\{c_i\} : c_i \in C_{missing}\}$

\STATE $S$ $\leftarrow$ $S \cup S_{manual}$ \\
\COMMENT{\textcolor{olive}{\textit{Step 3: fine-tuning}}}
\FOR{each epoch}
\STATE $S_{rand} \leftarrow$ randomly select $r$ percent of $S$
\STATE $Prompts, Labels \leftarrow \emptyset$
\FOR{each $S_i=\{h_1,...,h_n\}$ in $S_{rand}$}

\STATE $Table_{i}(t_1,...,t_n)$ $\leftarrow$ \textbf{PseudoTableGeneration}($S_i$, $P$, $k$)
\STATE $prompt_{i}$ $\leftarrow$\textbf{PromptConstruction}($Table_{i}$, $C_{learn}$)
\STATE Add $prompt_{i}$ to $Prompts$
\STATE Add $label_i=(h_1,..., h_n)$ to $Labels$

\ENDFOR

\FOR{each batch $(Prompts_\text{batch}, Labels_\text{batch})$}% in training data}
    \STATE $Outputs \leftarrow M_a(Prompts_\text{batch})$
    \STATE $Loss \leftarrow Loss(Outputs, Labels_\text{batch})$
    \STATE $M_a \leftarrow UpdateWeights(M_a, Loss)$
\ENDFOR
\ENDFOR
\RETURN $M_a$
\end{algorithmic}
\end{algorithm}
%\vspace{-5mm}

\section{ZTab}
\label{sec:ztab}
% ZTab works in two phases (Figure \ref{fig: overall ZTab}(b)). 

In \textbf{Learning Phase}, ZTab fine-tunes an annotation LLM $M_a$ using structured fine-tuning pseudo-tables constructed based on the \textit{class prototypes}  of the classes in $C_{learn}$ and the table schemas in $S$. For each class $c$ in $C_{learn}$, the class prototype of $c$ is a set of representative instances of $c$ generated using a prototype LLM $M_p$. The pseudo-tables serve as the training data to familiarize the annotation LLM with structured tabular data. 
By generating fine-tuning pseudo-data from leveraging the LLM $M_p$ pre-trained on the world corpus, ZTab addresses the data availability challenge and robustness across related domains and ontologies without retraining. In \textbf{Prediction Phase}, fine-tuned LLM $M_a$ is used to predict semantic types of columns in new tables.

We consider two variants of ZTab. \textbf{ZTab-privacy} uses open-source LLMs, avoiding privacy concerns related to sharing sensitive data with third parties. \textbf{ZTab-performance} uses more powerful closed-source LLMs, but learning and deployment are performed at the LLMs' owner sites. This variant is used when performance outweighs privacy concerns. The two variants differ in generating fine-tuning pseudo-tables due to different access levels to LLMs. We describe their learning procedures in Sections \textit{ZTab-privacy} and \textit{ZTab-performance}, then the prediction procedure in Section \textit{Prediction}.

\subsection{ZTab-privacy}

The learning phase, Algorithm \ref{alg:learning}, has three main steps:  

\textit{Class prototype generation:} For each semantic type \(c_i \in C_{learn}\), it queries the LLM $M_p$, the prototype LLM, to generate up to $e$ examples of \(c_i\) serving as its class prototype. The detail is captured by the function \textbf{ClassPrototypeGeneration}($c_i$, $e$, $M_p$) explained shortly. These class prototypes, denoted by $P$, are used to generate fine-tuning pseudo-tables during the fine-tuning step.

\textit{Handling missing classes:} 
This step creates one table schema for each class in $C_{learn}$ not contained in any table schema $S_i \in S$ and add these schemas to $S$, to ensure that all semantic types/classes in $C_{learn}$ are represented in the fine-tuning process.

\textit{Fine-tuning:} This step fine-tunes the annotation LLM $M_a$ in multiple epochs. During each epoch, it randomly selects $r$ percent of the table schemas from $S$, denoted $S_{rand}$, and 
% Redundant schemas are common in collections, so randomly selecting a portion for each epoch is a practical choice. 
constructs a pseudo-table $Table_i$ of $k$ rows for each schema \( S_i \) in $S_{rand}$, done by the function \textbf{PseudoTableGeneration}($S_i$, $P$, $k$). Note that the pseudo-table for $S_i$ is constructed ``on the fly" during each epoch; thus, it could be different for the same schema $S_i$ in a different epoch, increasing data diversity. It then represents $Table_i$ by creating one prompt for each column in the table, done by the function \textbf{PromptConstruction}($Table_{i}, C_{learn}$), and stores the prompts in $prompt_i$. The prompts $Prompt$ and the table headers $Labels$ for the schemas in $S_{rand}$ are used to fine-tune $M_a$ (i.e., lines 18-22), on a batch basis as in \cite{brown2020language}. 
The fine-tuning cost is determined by the schema sampling ratio $r$ and the row size $k$. 
A small value of $k$ is often sufficient for good performance of ZTab. More details are given in Section \ref{sec:ablation}. 

\textbf{ClassPrototypeGeneration($c_i$, $e$, $M_p$)}: 
For the class \( c_i \), the prototype LLM $M_p$ is provided with the prompt \textit{``Generate $e$ real-world examples of the semantic type $c_i$ commonly found in web tables."}, where \(e\) is the class prototype size. In response, $M_p$ generates up to $e$ instances for the class \( c_i \). For example, with $c_i$ being the class 'City' and $e=50$, $M_p$ will generate up to 50 city names to form the class prototype for the class City. 
%After generating descriptions for all classes, we obtain a complete set of descriptions \( D = \{d_1, d_2, \dots, d_m\} \), where each \( d_i \) corresponds to a distinct class in \( C_{learn} \). 
Note that $M_p$  can be replaced with any knowledge base, such as DBpedia or Wikidata; we choose an LLM pre-trained on the world corpus for better generalization.

\textbf{PseudoTableGeneration($S_i$, $P$, $k$)}: This function populates the table schema $S_i$ by randomly selecting $k$ values from the corresponding class prototype in $P$ for each semantic type in $S_i$. Thanks to the random selection of values from the class prototype, for the same schema $S_i$ selected in different epochs, the table generated for $S_i$ could be very different, allowing the fine-tuning to encounter a diverse range of training examples, which is essential for better model generalization. An alternative is to generate tables (one table at a time) using LLMs directly, but it produces low-quality tables as discussed in \cite{berkovitch2025generating}. Our fine-tuning benefits from a large number of diverse tables that are generated efficiently from class prototypes. 

\begin{figure}[t]
\centering
\begin{small}
\begin{minipage}{0.95\linewidth}
\textcolor{blue}{These are values of columns in a table. Each column starts with Column: followed by the values of that column. First, look at all the columns to understand the context of the table.}

\vspace{1mm}
\textcolor{purple}{Column 1: $t_{11}, t_{12}, \dots, t_{1k}$}

\textcolor{purple}{Column 2: $t_{21}, t_{22}, \dots, t_{2k}$}

\textcolor{purple}{...}

\textcolor{purple}{Column n: $t_{n1}, t_{n2}, \dots, t_{nk}$}

\vspace{1mm}
\textcolor{orange}{Your task is to annotate the Target Column using one semantic type that matches the values of the Target Column and the context of the table from the following list: $c_1, c_2, \dots, c_m$.}

\vspace{1mm}
\textcolor{teal}{Target Column: $t_{i1}, t_{i2}, \dots, t_{ik}$}

\textcolor{teal}{Semantic Type:}
\end{minipage}
\end{small}
\caption{Prompt for target column $t_i$ in a table with $n$ columns and $k$ rows.}
\label{quote:prompt_design}
\vspace{-5mm}
\end{figure}

\textbf{PromptConstruction ($Table$, $C$)}: 
For a given \(Table = (t_1, t_2, \dots, t_n)\) with \(n\) columns and \(k\) rows and a collection of  semantic types \(C = \{c_1, c_2, \dots, c_m\}\), this function generates $n$ prompts, one for each column in  $Table$. Figure \ref{quote:prompt_design} shows the prompt generated for a column \(t_{i} \) $\in Table$, which has four parts:
% \begin{enumerate}
(1) \textcolor{blue}{\textbf{Introduction}}: The general instruction about the table structure. (2) \textcolor{purple}{\textbf{Table Presentation}}: The table data presented column-by-column. (3) \textcolor{orange}{\textbf{Task Description}}: The instruction for the LLM to annotate the target column. (4) \textcolor{teal}{\textbf{Target Column}}: The target column for which the model is expected to predict the semantic type.
% \end{enumerate}
While each prompt focuses on prediction for one target column, the entire table data is presented in the prompt to help infer the target column's semantic type in the context of other columns in the table. 
This prompt design is one of four alternative designs, i.e., column-by-column table presentation and single column prediction, which allows the annotation LLM to benefit from the coherence within each column and focus on the prediction for one column at a time. Alternatively, the table can be presented row-by-row and the prediction can be made for all columns together, a common practice in the literature \cite{korini2023column, korini2024column}. We will study the effect of different prompt designs in Section \ref{sec:ablation}.

\subsection{ZTab-performance}
ZTab-performance follows the same overall design as ZTab-privacy but adapts its learning procedure to comply with the closed-source fine-tuning policy (e.g.,  OpenAI fine-tuning policy), which requires providing a fixed training dataset in advance of fine-tuning, rather than generating new samples during fine-tuning like Algorithm 1. 
Accordingly, all pseudo-tables are pre-generated prior to submission for fine-tuning, as illustrated in Figure 1(b). While this setup allows ZTab-performance to leverage large closed-source models for higher accuracy, it limits the diversity of training data compared to ZTab-privacy, where pseudo-tables are dynamically constructed in each epoch. 
In essence, ZTab-performance trades off some data diversity for compatibility with managed OpenAI fine-tuning and the performance advantages of larger proprietary models.

\begin{algorithm}[t]
\small
\caption{Prediction Phase of ZTab}
\label{alg:prediction}
\begin{algorithmic}[1]
\REQUIRE New Table $T = (t_1, t_2, \dots, t_n)$, set of semantic types/classes $C_{pred}$, Fine-tuned Model $M_a$ from Algorithm \ref{alg:learning}
\ENSURE Predicted Class for Each Column in $T$

\STATE $Prompts \leftarrow$\textbf{PromptConstruction}($T$, $C_{pred}$)

\FOR{each $prompt_{i}$ in $Prompts$ corresponding to column $t_i$}
    \STATE $Output_{i} \leftarrow M_a(prompt_{i})$ \\
    \COMMENT{\textcolor{olive}{\textit{label remapping}}}    
    \STATE $h_i \leftarrow \arg\max\limits_{c_j \in C_{pred}} \text{cosine\_sim}(E(M_a,Output_{i}), E(M_a,c_j))$
\ENDFOR
\RETURN $(h_1, ..., h_n)$

\end{algorithmic}
\end{algorithm}

\subsection{Prediction}
The fine-tuned model $M_a$ produced by 
Algorithm \ref{alg:learning} is applied to annotate the columns in a test table $T$ without headers using the semantic types in \( C_{pred}\). Algorithm \ref{alg:prediction} presents this phase.  
First, \textbf{PromptConstruction ($T$, $C_{pred}$)} is used to generate the prompts for the columns in $T$ but using the classes $C_{pred}$ instead of $C_{learn}$.
% The same prompt format used during the learning phase (Figure \ref{quote:prompt_design}) is employed here, allowing the model to generalize learned associations to new tables. 
$M_a(prompt_i)$ denotes the tokens generated by $M_a$ for $prompt_i$, which may not be a class in \( C_{pred}\). 
\textit{label remapping} is used to map the generated token to the most similar class in \( C_{pred} \), i.e.,  $h_i$, where similarity is measured based on the embeddings $E(M_a,\cdot)$ extracted from $M_a$.
Finally, $(h_1,\cdots,h_n)$ are returned as the predicted semantic types for the columns $(t_1,\cdots,t_n)$.
% We measure the similarity based on the embedding \( E(M, x) \) of the entire string \( x \), extracted using the fine-tuned model \( M \). 
The distinction of In-Domain, Cross-Domain, and Cross-Ontology Generalizations lies in the class composition in \( C_{pred}\) and the value composition in $T$; no additional change is needed.

%\textcolor{brown}{
\subsection{Extensibility to Complex Table Structures}
\label{subsec:extensibility}
ZTab focuses on simple relational tables. There are two ways to extend ZTab to deal with more complex tables, e.g., merged cells, multi-level headers, or nested JSON.
(1) Preprocessing. We can first convert the input into the standard relational format that ZTab expects. For tables in templated documents (e.g., invoices/forms), we can extract relational tables using a document-to-structure system such as TWIX \cite{lin2025twix}.
For nested JSON, we can flatten nested fields into columns using key paths (e.g., \texttt{"Address": \{"City": ...\}} $\rightarrow$ column \texttt{Address.City})
\cite{discala2016automatic,bahta2019translating}.
For merged cells (\texttt{rowspan}/\texttt{colspan}), we can copy the merged value into all cells it covers \cite{chen2000mining,adelfio2013schema}.
(2) Framework adaptation. The second way is to modify ZTab to handle complex tables directly. In particular, this requires modifying \textbf{PseudoTableGeneration} to generate complex tables and modifying \textbf{PromptConstruction} to serialize such tables for the LLM. The first approach is preferred as it requires no change to ZTab.     
%}

\section{Evaluation}
We evaluate the performance of ZTab in the three scenarios of In-Domain, Cross-Domain, and Cross-Ontology Generalization. All experiments were conducted with 4 NVIDIA RTX 6000 Ada. The source code is available at \href{https://github.com/hoseinzadeehsan/ZTab}{https://github.com/hoseinzadeehsan/ZTab}.

\subsection{Evaluation Method}

%\begin{comment}
\begin{table}[h]
\vspace{-3mm}
\centering
\caption{Summary of datasets}
\label{table:dataset}
\vspace{-2mm}
\begin{tabular}{l|ccc}
Dataset & \# Class & \# Training Tables & \# Test Tables\\
\hline
WikiTable  & $255$ & 397098 & 4764\\
SOTAB\textsubscript{sch}  & $82$ & $44769$ & 609  \\
SOTAB\textsubscript{sch-s}  & $82$ & $10631$ & 609\\
SOTAB\textsubscript{dbp}  & $46$ & $37631$ & 279\\
T2D  & $37$ & $160$ & 109\\
Efthymiou  & $31$ & - & 614\\
Limaye  & $8$ & - & 114\\
\end{tabular}
\vspace{-2mm}
\end{table}
%\end{comment}
\subsubsection{\textbf{Datasets}}
\label{sec:dataset}
We evaluate ZTab on seven datasets: WikiTable \cite{deng2020turl}, T2D, Limaye, Efthymiou \cite{chen2019learning}, and three datasets from SOTAB-V2, i.e., SOTAB\textsubscript{sch}, SOTAB\textsubscript{sch-s}, and SOTAB\textsubscript{dbp} \cite{sotab}, summarized in Table~\ref{table:dataset}. Limaye and Efthymiou have only test tables. Instead of the older SOTAB-91 and SOTAB-27 benchmarks used in \cite{feuer2024archetype}, we adopt the updated SOTAB-V2 suite, which includes manually verified validation and test splits and provides both schema.org and DBpedia annotations, offering a more robust benchmark for LLM-based models. %We exclude Wikitable \cite{deng2020turl} as it is a multi-label dataset, whereas ZTab is designed for single-label classification. 
Wikitable is a multi-label dataset and we select the first available label when multiple labels are present. 
VizNet \cite{zhang2019sato,suhara2022annotating} is excluded due to significant label overlap and noise \cite{ babamahmoudi2025evaluating}. 

First, we discuss the construction of the input components $S$, $C_{learn}$, and $C_{pred}$ for the domain-based zero-shot column annotation problem. Note that these input components do not utilize the actual tabular content in these datasets, as required by the zero-shot requirement.

\textit{In-Domain Generalization:} For each dataset, 
% the training and testing data belong to the same domain and ontology, which is ideal for In-Domain Generalization. 
$S$ is extracted by including table header of each training table with duplicates preserved. $C_{learn}$ and $C_{pred}$ are set to class list of dataset.

\textit{Cross-Domain Generalization:} 
% T2D is sourced from the web while Limaye and Efthymiou originate from Wikipedia. 
Following the settings in \cite{chen2019learning}, Limaye and Efthymiou are annotated using a subset of the classes from T2D, making them ideal for the Cross-Domain Generalization. We extract $S$ and $C_{learn}$ from T2D as the learning domain and evaluate the fine-tuned model on Limaye and Efthymiou as two test datasets. We set $C_{pred}$ to the classes of the test dataset involved. 

\textit{Cross-Ontology Generalization:} SOTAB\textsubscript{dbp}, SOTAB\textsubscript{sch}, and SOTAB\textsubscript{sch-s} datasets are derived from the general web domain but represent two distinct ontologies: Schema.org (denoted by sch, sch-s) and DBpedia (denoted by dbp). This dual-ontology design makes them suitable for evaluating the Cross-Ontology Generalization scenario. 
For this scenario, we extract $S$ and $C_{learn}$ from SOTAB\textsubscript{sch} and SOTAB\textsubscript{sch-s} as two domains for fine-tuning and evaluate the fine-tuned models on the test data SOTAB\textsubscript{dbp}. $C_{pred}$ is set to the classes of the test dataset.

\subsubsection{\textbf{Metrics and Algorithms Evaluated}}
\label{sec:algorithms}
We measure the \textit{micro F1-score} collected on test tables, following previous works \cite{feuer2024archetype,korini2023column, korini2024column,hoseinzadeefficient}. 
All F1-score values are multiplied by 100 (e.g., 80\% is written as 80).
We evaluate ZTab against baseline and reference algorithms detailed as follows:

\textbf{ZTab-privacy, denoted by ZTab-pri($M_p,M_a$):} We use the open-source Llama3.1-70B \cite{dubey2024llama} as the prototype LLM $M_p$ and use the open-source Qwen2.5-Coder-1.5B \cite{hui2024qwen2}, Llama-7B \cite{touvron2023llama}, Mistral-7B \cite{jiang2023mistral}, and Qwen2.5-7B \cite{yang2025qwen3} as the annotation LLM $M_a$. 
Fine-tuning of Qwen2.5-7B is done with LoRA \cite{hulora} (rank = 256) using \textit{transformers} and \textit{peft} libraries, batch size of 1, gradient accumulation of 8 for memory efficiency, learning rate of \(1 \times 10^{-5}\).
The epoch number is set to 20. For datasets with validation data (SOTAB), we evaluate the best model on validation performance and for datasets without validation data (T2D, Limaye, and Efthymiou) we evaluate the model after the final epoch. The class prototype size \( e \) is set to 500, and the schema sampling rate $r$ is set to 2.5\% for the SOTAB datasets, 0.5\% for WikiTable, and to 100\% for T2D due to its small size. Row size \( k \) is set to 3. Unless otherwise stated, these settings remain consistent for ZTab-privacy. 

\textbf{ZTab-performance, denoted by ZTab-per($M_p,M_a$):} Both prototype LLM ($M_p$) and annotation LLM ($M_a$) are from the same GPT models: GPT-3.5, GPT-4o-mini, GPT-4o, and GPT-4.1-mini. %We use the close-sources GPT-3.5, GPT-4o-mini, GPT-4o, and GPT-4.1-mini for both the prototype LLM $M_p$ and the annotation LLM $M_a$. 
A schema sampling ratio $r$ of 2.5\% is used for the SOTAB datasets, 0.5\% for WikiTable, and 100\% T2D due to its small size. Row size \( k \) is set to 3. Unless otherwise stated, these settings remain consistent for ZTab-performance.

\begin{table}[h]
\centering
\vspace{-3mm}
%\resizebox{\textwidth}{!}{%
%\begin{tabular}{@{}lp{5.2cm}ll@{}}
\caption{ZTab, baselines, and reference algorithms. Yes/No indicates whether a model uses training data and whether user information is disclosed at inference time.}
\label{tab:baseline}
\vspace{-2mm}
\resizebox{\columnwidth}{!}{%
\begin{tabular}{@{}lp{4.5cm}ll@{}}
\textbf{Category} & \textbf{Model} & \shortstack{\textbf{Train}\\\textbf{Data}} & \shortstack{\textbf{Inf.}\\\textbf{Disclosure}} \\
\midrule
\multirow{3}{*}{\shortstack[l]{\textbf{Ours}\\{\raggedright(Privacy)}}} 
% & \textcolor{brown}{ZTab-pri (Llama3.1-70B,Qwen-1.5B)} & \textcolor{brown}{No} & \textcolor{brown}{No} \\
& ZTab-pri (Llama3.1-70B,Qwen-1.5B) & No & No \\
& ZTab-pri (Llama3.1-70B,Llama-7B) & No & No \\
& ZTab-pri (Llama3.1-70B,Mistral-7B) & No & No \\
& ZTab-pri (Llama3.1-70B,Qwen2.5-7B) & No & No \\
\cmidrule(lr){2-4}
\multirow{4}{*}{\shortstack[l]{\textbf{Ours}\\{\raggedright(Performance)}}} 
& ZTab-per (GPT-4.1-mini,GPT-4.1-mini) & No & Yes \\
& ZTab-per (GPT-4o-mini,GPT-4o-mini) & No & Yes \\
& ZTab-per (GPT-4o,GPT-4o) & No & Yes \\
& ZTab-per (GPT-3.5,GPT-3.5) & No & Yes \\
\midrule
\multirow{6}{*}{\shortstack[l]{\textbf{Baseline}\\{\raggedright(Privacy)}}} 
& TableLlama (Llama-7B) \cite{zhang2024tablellama} & No & No \\
& Jellyfish (Mistral-7B) \cite{zhang2024jellyfish} & No & No \\
& ArcheType\textsubscript{ZS} (T5) \cite{feuer2024archetype} & No & No \\
& GPT-20B-based \cite{korini2023column} & No & No \\
& GPT-120B-based \cite{korini2023column} & No & No \\
& Llama3.1-70B-based \cite{korini2023column} & No & No \\
\cmidrule(lr){2-4}
\multirow{6}{*}{\shortstack[l]{\textbf{Baseline}\\{\raggedright(Performance)}}} 
& GPT-3.5-based \cite{korini2023column} & No & Yes \\
& Chorus (GPT-3.5) \cite{kayali2024chorus} & No & Yes \\
& CENTS (GPT-3.5) \cite{xiao2025cents} & No & Yes \\
& CENTS (GPT-4o-mini) \cite{xiao2025cents} & No & Yes \\
& CENTS (GPT-4.1-mini) \cite{xiao2025cents} & No & Yes \\
& CENTS (GPT-4o) \cite{xiao2025cents} & No & Yes \\
\midrule
\multirow{2}{*}{\shortstack[l]{\textbf{Reference}\\{\raggedright(Supervised)}}} 
& Doduo (BERT) \cite{suhara2022annotating} & Yes & No \\
& ArcheType\textsubscript{FT} (Llama-7B) \cite{feuer2024archetype} & Yes & No \\
\end{tabular}%
}

\vspace{-5mm}
\end{table}

%\textcolor{brown}{
\textbf{Baselines and Reference Upper Bounds.}
Table~\ref{tab:baseline} groups prior methods into (i) \emph{baselines} that satisfy our deployment constraint, \textbf{no access to user-specific labeled tables}, and (ii) \emph{reference} that \textbf{do} train on user-provided labeled tables.
ZTab belongs to the first group: it is fine-tuned only from \emph{schema-level configuration} (a class list and schemas) and does not use any user labeled tables from the target deployment.
Therefore, we treat all methods that operate under the same no user labeled table constraint as baselines.
We additionally report supervised reference models to quantify the remaining \emph{gap-to-supervised} when user-provided labeled tables are available; these numbers should be interpreted as upper bounds rather than directly comparable competitors.
%}

For all baseline and reference models, the default settings reported in their papers are applied. We did not include supervised models like Sherlock \cite{hul2019sherlock}, SATO \cite{zhang2019sato}, TURL \cite{deng2020turl}, KGLink \cite{10597678}, TCN \cite{wang2021tcn}, RECA \cite{sun2023reca}, GAIT \cite{hoseinzade2024graph}, TABBIE \cite{iida2021tabbie}, ColNet \cite{chen2019colnet}, HNN \cite{chen2019learning} because Doduo and ArcheType are two of the best performing supervised models \cite{feuer2024archetype,suhara2022annotating}. 
The columns ``Train data" and ``Inf. Disclosure" in Table \ref{tab:baseline} indicate whether user-specific training data is used (like all supervised models) and whether user data is sent to a third party at inference time (like all models relying on a closed-source LLM). Data privacy can be at risk when either is Yes. ZTab-privacy and zero-shot baselines based on open-source LLMs, marked by (Privacy), address this privacy concern. 

\begin{table*}[t]
\vspace{-3mm}
\caption{Main results: we compare ZTab-privacy with zero-shot baselines that use open-source LLMs, and compare ZTab-performance with zero-shot baselines that use closed-source LLMs.}
\label{tab:model_comparison}
\vspace{-2mm}
\centering
\small
\setlength{\tabcolsep}{4pt}
\renewcommand{\arraystretch}{1.05}
\begin{tabular}{@{}llccc@{}}
\textbf{Category} & \textbf{Model} & \textbf{In-Domain} & \textbf{Cross-Domain} & \textbf{Cross-Ontology} \\
\midrule

% ---------- ZTab-Privacy Comparison ----------

\multirow[t]{14}{*}{%
  \begin{minipage}[t]{3.8cm} % adjust width to fit your first column
    \raggedright
    \rule{0pt}{1.2\baselineskip}% top strut to align with the row baseline
    \textbf{Comparing ZTab-privacy }\\
    \textbf{with zero-shot baselines }\\
    \textbf{that use open-source LLMs}%
  \end{minipage}%
}

%\cmidrule(lr){2-5}
& TableLlama (Llama-7B)                        & 48.2 & 77.5 & 38.1 \\
& \textbf{ZTab-pri (Llama3.1-70B,Llama-7B)}      & \textbf{65.7} & \textbf{82.0} & \textbf{67.1} \\
\cmidrule(lr){2-5}
& Jellyfish (Mistral-7B)                       & 24.0 & 68.5 & 26.7 \\
& \textbf{ZTab-pri (Llama3.1-70B,Mistral-7B)}    & \textbf{69.1} & \textbf{87.4} & \textbf{72.1} \\
\cmidrule(lr){2-5}
& ArcheType\textsubscript{ZS} (T5)             & 43.3 & 73.0 & 48.3 \\
& GPT-20B-based                       & 21.9 & 80.0 & 26.6 \\
& GPT-120B-based                      & 8.0 & 47.6 & 15.3 \\
& Llama3.1-70B-based                  & 46.3 & 89.3 & 64.2 \\
& \textbf{ZTab-pri (Llama3.1-70B,Qwen-1.5B)}      & \textbf{65.0} & \textbf{81.7} & \textbf{68.1} \\
% & \textcolor{brown}{\textbf{ZTab-pri (Llama3.1-70B,Qwen-1.5B)}}      & \textcolor{brown}{\textbf{65.0}} & \textcolor{brown}{\textbf{81.7}} & \textcolor{brown}{\textbf{68.1}} \\
& \textbf{ZTab-pri (Llama3.1-70B,Qwen2.5-7B)}      & \textbf{71.7} & \textbf{90.7} & \textbf{73.7} \\

\midrule
% ---------- ZTab-Performance Comparison ----------

\multirow[t]{14}{*}{%
  \begin{minipage}[t]{3.8cm} % adjust width to fit your first column
    \raggedright
    \rule{0pt}{1.2\baselineskip}% top strut to align with the row baseline
    \textbf{Comparing ZTab-performance }\\
    \textbf{with zero-shot baselines }\\
    \textbf{that use closed-source LLMs}%
  \end{minipage}%
}

& Chorus (GPT-3.5)                            & 56.8 & 85.0 & 64.8 \\
& GPT-3.5-based                       & 63.1 & 86.6 & 70.0 \\
& CENTS (GPT-3.5)                              & 53.3 & 88.7 & 53.7 \\
& \textbf{ZTab-per (GPT-3.5,GPT-3.5)}                      & \textbf{74.0} & \textbf{91.9} & \textbf{76.9} \\
\cmidrule(lr){2-5}
& CENTS (GPT-4o-mini)                          & 61.2 & 86.2 & 61.1 \\
& \textbf{ZTab-per (GPT-4o-mini,GPT-4o-mini)}                  & \textbf{74.6} & \textbf{93.4} & \textbf{77.2} \\
\cmidrule(lr){2-5}
& CENTS (GPT-4o)                               & 72.2 & 91.7 & 74.5 \\
& \textbf{ZTab-per (GPT-4o,GPT-4o)}                       & \textbf{76.6} & \textbf{94.4} & \textbf{78.3} \\
\cmidrule(lr){2-5}
& CENTS (GPT-4.1-mini)                         & 72.1 & 83.0 & 73.6 \\
& \textbf{ZTab-per (GPT-4.1-mini,GPT-4.1-mini)}                 & \textbf{76.7} & \textbf{94.3} & \textbf{78.3} \\

\midrule
% ---------- Supervised (Reference) ----------

\multirow{2}{*}{\shortstack{\textbf{Supervised}\\(Reference)}} 
& Doduo (BERT)                                 & 83.8 & 63.9 & --- \\
& ArcheType\textsubscript{FT} (Llama-7B)       & 83.3 & 55.3 & --- \\

\end{tabular}

\vspace{-5mm}
\end{table*}

\subsection{Main Results}
\label{mainresults}

Table \ref{tab:model_comparison} summarizes the comparison results on micro-F1 score, averaged over applicable datasets, under In-Domain, Cross-Domain, and Cross-Ontology Generalization. The results are grouped separately for ZTab-privacy and ZTab-performance. The former considers the baselines that use open-source LLMs and the latter considers the baselines that use closed-source LLMs. For a fair comparison with baselines built on different GPT versions, we include corresponding ZTab-performance with annotation LLMs based on the same GPT models. The winners are highlighted in boldface. 
Below, we explain the key findings here and provide the detailed analysis in Sections \ref{sec:privacy} and \ref{sec:performnace}. 

\textbf{In-Domain Generalization}: The average performance is reported using all datasets except for Limaye and Efthymiou, which do not have the training data required for supervised models. ZTab-privacy outperforms the best zero-shot model, TableLlama, by 23.5\% and ZTab-performance outperforms the best zero-shot model, CENTS (GPT-4o), by 4.5\%. 

\textbf{Cross-Domain Generalization:} ZTab-privacy 
outperforms the best open-source zero-shot model, LLM-ZeroShot (Llama3.1-70B), by 1.4\%, and ZTab-performance outperforms the best closed-source zero-shot model, CENTS (GPT-4o), by 2.7\%, thanks to the fine-tuning of ZTab. 

\textbf{Cross-Ontology Generalization:} The results demonstrate ZTab’s robustness of generalizing across different ontologies. ZTab-privacy outperforms the best open-source zero-shot baseline, LLM-ZeroShot (Llama3.1-70B), by 9.5\% and ZTab-performance outperforms the best closed-source zero-shot model, CENTS(GPT-4o), by 3.8\%. 

Besides the above comparison with the baselines, we would like to highlight several other properties of ZTab below.

\textbf{Privacy vs Performance:} Closed-source models achieve higher accuracy but require data sharing, limiting their use in privacy-sensitive settings. Open-source models ensure full privacy but perform worse overall. ZTab-privacy bridges this gap, offering competitive results while preserving privacy, whereas ZTab-performance achieves the best accuracy when privacy is not a concern.

%\textcolor{brown}{
\textbf{Deployment Effectiveness:} ZTab-privacy with the compact Qwen-1.5B achieves strong performance in compute-constrained settings. Despite being $\sim$47$\times$ smaller than the strongest open-source zero-shot baseline (Llama3.1-70B), it outperforms it in In-Domain (+18.7\%) and Cross-Ontology (+3.9\%) settings. Compared with ZTab-privacy using the larger Qwen2.5-7B annotator, the 1.5B variant incurs a modest drop of 6.7\% (In-Domain), 9.0\% (Cross-Domain), and 5.6\% (Cross-Ontology) micro-F1, indicating that ZTab remains effective with a small deployable annotation model.

%\textcolor{brown}{
\textbf{ZTab vs Supervised (reference) models:}
In \textit{In-Domain Generalization}, supervised models naturally outperform ZTab due to their access to complete user-provided training data, exceeding ZTab-performance by margins of 7.1\% (Doduo) and 6.6\% (ArcheType). However, we should mention that supervised models are not an option under our zero-shot constraint, where no user-specific labeled training data is available; they are included only to measure the performance gap of ZTab due to meeting this constraint.
In \textit{Cross-Domain Generalization}, ZTab-privacy and ZTab-performance outperform the best supervised model, Doduo, by 26.8\% and 30.5\%, respectively. By leveraging class prototypes from a pre-trained LLM, ZTab learns from pseudo-tables that are free from domain-specific biases, enhancing generalization on Efthymiou and Limaye. In contrast, supervised models trained on T2D perform poorly on these datasets due to over-fitting on domain-specific data.
Finally, supervised models are inapplicable in \textit{Cross-Ontology Generalization} as they rely on fixed label sets and cannot adapt to new ontologies without retraining.

\begin{table*}[t]
\vspace{-3mm}
\caption{Details of Comparing ZTab-privacy with open-source based models for In-domain generalization.}
\label{table:in_domain}
\vspace{-2mm}
\centering
\begin{tabular}{@{}lllllllllll@{}}
 &  & 
\multicolumn{7}{c}{\textbf{In-Domain}} \\ 
\cmidrule(lr){3-9} 
\textbf{Category} & \textbf{Model} & 
\textbf{WikiTable} & \textbf{SOTAB\textsubscript{sch}} & 
\textbf{SOTAB\textsubscript{sch-s}} & \textbf{SOTAB\textsubscript{dbp}} & 
\textbf{T2D} & \textbf{Efthymiou} & \textbf{Limaye} \\

\midrule
\multirow{9}{*}{Zero-shot}

 %&\\textbf{ZTab-pri (Llama3.1-70B,Qwen-1.5B)} &  & \textbf{} & \textbf{} & \textbf{} & \textbf{} & \textbf{} & \textbf{} \\
%\cmidrule(l){2-9}
 & TableLlama (Llama-7B) & 59.6 & 32.6 & 32.6 & 38.1 & 78.2 & 68.9 & 86.0 \\
 & \textbf{ZTab-pri (Llama3.1-70B,Llama-7B)} & 24.7 & \textbf{73.1} & \textbf{71.9} & \textbf{69.6} & \textbf{90.4} & \textbf{82.5} & \textbf{85.9} \\
\cmidrule(l){2-9}
 & Jellyfish (Mistral-7B) & 0.0 & 8.4 & 8.4 & 26.7 & 76.7 & 58.9 & 78.0 \\
 & \textbf{ZTab-pri (Llama3.1-70B,Mistral-7B)} & \textbf{30.6} & \textbf{74.6} & \textbf{73.5} & \textbf{73.2} & \textbf{93.5} & \textbf{87.0} & \textbf{91.1} \\
\cmidrule(l){2-9}
 & ArcheType\textsubscript{ZS} (T5) & 6.3 & 40.1 & 40.1 & 48.3 & 81.6 & 67.7 & 78.3 \\
 & GPT-20B-based & 3.4 & 7.8 & 7.8 & 26.6 & 64.0 & 73.1 & 86.8 \\
 & GPT-120B-based & 1.2 & 4.2 & 4.2 & 15.3 & 15.1 & 19.7 & 75.4 \\
 & Llama3.1-70B-based & 7.5 & 32.6 & 32.6 & 64.2 & 94.7 & 87.3 & 91.2 \\
  & \textbf{ZTab-pri (Llama3.1-70B,Qwen-1.5B)} & \textbf{19.8} & \textbf{73.6} & \textbf{69.3} & \textbf{70.1} & 92.0 & 76.8 & 85.9 \\
  %& \textcolor{brown}{\textbf{ZTab-pri (Llama3.1-70B,Qwen-1.5B)}} & \textcolor{brown}{\textbf{19.8}} & \textcolor{brown}{\textbf{73.6}} & \textcolor{brown}{\textbf{69.3}} & \textcolor{brown}{\textbf{70.1}} & \textcolor{brown}{92.0} & \textcolor{brown}{76.8} & \textcolor{brown}{85.9} \\
 & \textbf{ZTab-pri (Llama3.1-70B,Qwen2.5-7B)} & \textbf{34.1} & \textbf{76.9} & \textbf{75.1} & \textbf{76.2} & \textbf{96.2} & \textbf{91.3} & \textbf{92.5} \\
 %& \textbf{ZTab-universal (Qwen2.5-7B)} & \textbf{} & \textbf{} & \textbf{} & \textbf{} & \textbf{} & \textbf{} & \textbf{} \\
\midrule
% \multirow{2}{*}{\textcolor{brown}{\shortstack{Supervised\\(Reference)}}} 
%  & \textcolor{brown}{Doduo (BERT)} & \textcolor{brown}{75.2} & \textcolor{brown}{86.3} & \textcolor{brown}{81.1} & \textcolor{brown}{85.2} & \textcolor{brown}{91.1} & \textcolor{brown}{-} & \textcolor{brown}{-} \\
%  & \textcolor{brown}{ArcheType\textsubscript{FT} (Llama-7B)} & \textcolor{brown}{76.7} & \textcolor{brown}{85.1} & \textcolor{brown}{83.0} & \textcolor{brown}{83.6} & \textcolor{brown}{88.0} & \textcolor{brown}{-} & \textcolor{brown}{-} \\
\multirow{2}{*}{\shortstack{Supervised\\(Reference)}} 
 & Doduo (BERT) & 75.2 & 86.3 & 81.1 & 85.2 & 91.1 & - & - \\
 & ArcheType\textsubscript{FT} (Llama-7B) & 76.7 & 85.1 & 83.0 & 83.6 & 88.0 & - & - \\
\end{tabular}

\vspace{5mm}

\centering
\caption{Details of comparing ZTab-privacy with open-source based models for Cross-domain (left) and cross-ontology generalization (right). 
% Each ZTab variant corresponds to the same backbone model (Llama-7B, Mistral-7B, or Qwen2.5-7B). 
Supervised models are not applicable in cross-ontology generalization.}
\label{table:generalization}
\vspace{-2mm}
%\resizebox{\textwidth}{!}{%
\begin{tabular}{@{}l l  ccc | ccc@{}}
%\toprule
 &  & 
\multicolumn{3}{c}{\textbf{Cross-Domain (Test Datasets)}} & 
\multicolumn{3}{c}{\textbf{Cross-Ontology (Learning Datasets)}} \\
\cmidrule(lr){3-5} \cmidrule(lr){6-8}
\textbf{Category} & \textbf{Model} & 
\textbf{T2D} & \textbf{Efthymiou} & \textbf{Limaye} & 
\textbf{SOTAB\textsubscript{dbp}} & \textbf{SOTAB\textsubscript{sch}} & \textbf{SOTAB\textsubscript{sch-s}} \\
\midrule

\multirow{8}{*}{Zero-shot}

 %&\\textbf{ZTab-pri (Llama3.1-70B,Qwen-1.5B)} &  & \textbf{} & \textbf{} & \textbf{} & \textbf{} & \textbf{} & \textbf{} \\
%\cmidrule(l){2-8}

 & TableLlama (Llama-7B) & 78.2 & 68.9 & 86.0 & 38.1 & 38.1 & 38.1 \\
  & \textbf{ZTab-pri (Llama3.1-70B,Llama-7B)} & \textbf{90.4} & \textbf{80.3} & \textbf{83.6} & \textbf{69.6} & \textbf{67.4} & \textbf{66.7} \\
\cmidrule(l){2-8}
 & Jellyfish (Mistral-7B) & 76.7 & 58.9 & 78.0 & 26.7 & 26.7 & 26.7 \\
 & \textbf{ZTab-pri (Llama3.1-70B,Mistral-7B)} & \textbf{93.5} & \textbf{85.4} & \textbf{89.3} & \textbf{73.2} & \textbf{72.5} & \textbf{71.6} \\
\cmidrule(l){2-8}
 & ArcheType\textsubscript{ZS} (T5) & 81.6 & 67.7 & 78.3 & 48.3 & 48.3 & 48.3 \\
 & GPT-20B-based & 64.0 & 73.1 & 86.6 & 26.6 & 26.6 & 26.6 \\
 & GPT-120B-based & 15.1 & 19.7 & 75.4 & 15.3 & 15.3 & 15.3 \\
 & Llama3.1-70B-based & 94.7 & 87.3 & 91.2 & 64.2 & 64.2 & 64.2 \\
 %& TabGen & 90.6 & 63.4 & 82.3 & 47.7 & 39.6 & 37.1 \\
& \textbf{ZTab-pri (Llama3.1-70B,Qwen-1.5B)} & 92.0 & 78.6 & 84.7 & \textbf{70.1} & \textbf{68.7} & \textbf{67.5}  \\
%& \textcolor{brown}{\textbf{ZTab-pri (Llama3.1-70B,Qwen-1.5B)}} & \textcolor{brown}{92.0} & \textcolor{brown}{78.6} & \textcolor{brown}{84.7} & \textcolor{brown}{\textbf{70.1}} & \textcolor{brown}{\textbf{68.7}} & \textcolor{brown}{\textbf{67.5}}  \\
 & \textbf{ZTab-pri (Llama3.1-70B,Qwen2.5-7B)} & \textbf{96.2} & \textbf{88.9} & \textbf{92.4} & \textbf{76.2} & \textbf{74.2} & \textbf{73.1} \\
 %& \textbf{ZTab-universal (Qwen2.5-7B)} & \textbf{} & \textbf{} & \textbf{} & \textbf{} & \textbf{} & \textbf{} \\
\midrule
% \multirow{4}{*}{\textcolor{brown}{\shortstack{Supervised\\(Reference)}}}
%  & \textcolor{brown}{Doduo (BERT)} & \textcolor{brown}{91.1} & \textcolor{brown}{63.2} & \textcolor{brown}{64.5} & \textcolor{brown}{--} & \textcolor{brown}{--} & \textcolor{brown}{--} \\
%  & \textcolor{brown}{ArcheType\textsubscript{FT} (Llama-7B)} & \textcolor{brown}{88.0} & \textcolor{brown}{53.0} & \textcolor{brown}{57.6} & \textcolor{brown}{--} & \textcolor{brown}{--} & \textcolor{brown}{--} \\
%  & \textcolor{brown}{ColNet (Deep learning)} & \textcolor{brown}{94.7} & \textcolor{brown}{61.9} & \textcolor{brown}{59.7} & \textcolor{brown}{--} & \textcolor{brown}{--} & \textcolor{brown}{--} \\
%  & \textcolor{brown}{HNN (Deep learning)} & \textcolor{brown}{96.6} & \textcolor{brown}{65.0} & \textcolor{brown}{74.6} & \textcolor{brown}{--} & \textcolor{brown}{--} & \textcolor{brown}{--} \\
\multirow{4}{*}{\shortstack{Supervised\\(Reference)}}
 & Doduo (BERT) & 91.1 & 63.2 & 64.5 & -- & -- & -- \\
 & ArcheType\textsubscript{FT} (Llama-7B)& 88.0 & 53.0 & 57.6 & -- & -- & -- \\
 & ColNet (Deep learning) & 94.7 & 61.9 & 59.7 & -- & -- & -- \\
 & HNN (Deep learning)& 96.6 & 65.0 & 74.6 & -- & -- & -- \\
%\bottomrule
\end{tabular}%
%}

\vspace{-5mm}
\end{table*}

\subsection{ZTab-privacy}
\label{sec:privacy}
%\vspace{-1mm}
This section provides the data-specific comparison for the summary results for ZTab-privacy in Table III. 
All models used here are fully open-source and can be run locally without sharing data with third parties. For a fair comparison with TableLlama (Llama-7B) and Jellyfish (Mistral-7B), which use LLMs of similar size, we consider ZTab-privacy with the annotation LLMs based on Llama-7B and Mistral-7B.%, while excluding the T5 variant since CENTS already outperforms ArcheType, making comparison with CENTS sufficient.

\subsubsection{In-Domain Generalization}
%\label{sec:in_domain}
Table \ref{table:in_domain} details the comparison on individual datasets. 
Note that supervised learning models cannot be applied to Limaye and Efthymiou datasets that do not have training data. TableLlama performs best on the WikiTable dataset because it was fine-tuned on this dataset’s training data.
For datasets in the left to right order in the table, 
ZTab (Llama3.1-70B,Qwen2.5-7B) outperforms the best zero-shot baseline
by 26.6\%, 44.3\%, 42.5\%, 12\%, 1.5\%, 4\% and 1.3\%, respectively.
TableLlama was fine-tuned using the training data of WikiTable, so we don't consider TableLlama as a zero-shot model for the WikiTable dataset. Roughly, the improvement increases as the number of classes in the dataset increases, highlighting a common difficulty of existing LLM based zero-shot models in dealing with a large number of classes where semantically similar classes can confuse the model. ZTab addresses this ambiguity by providing a few examples of each class via class prototypes, improving accuracy.

Supervised models achieve the highest micro-F1 due to their access to user-provided training data. Prior research has shown a significant overlap between the training and test data in the SOTAB datasets \cite{babamahmoudi2025evaluating}, which further enhances their results. However, these models are sensitive to the size of available data, as shown by the performance decline in the smaller SOTAB\textsubscript{sch-s}. In the small T2D, ZTab even surpasses these supervised models of Doduo and ArcheType by $5.1\%$, $8.2\%$ correspondingly. Last but not least, supervised models are not an option when labeled training data is not available or providing such data raises privacy concerns. In such cases, 
% ZTab is highly competitive considering its performance without training data. 
ZTab bridges the performance gap between zero-shot and supervised learning models while addressing data availability and privacy challenges.

\subsubsection{Cross-Domain Generalization}
%\label{sec:cross_domain}

Table \ref{table:generalization} (left) shows the results for Cross-Domain generalization, where models are trained on the domain of T2D and tested on T2D, Limaye, and Efthymiou. Two supervised models, HNN \cite{chen2019learning} and ColNet \cite{chen2019colnet}, are included due to their relevance to these datasets.  The zero-shot baseline models rely solely on the test dataset without fine-tuning, resulting in identical performance across all three generalization scenarios. 

ZTab performs consistently well on both test datasets Limaye and Efthymiou, demonstrating robustness to domain shifts. The use of class prototypes in ZTab allows the fine-tuning to be free of the biases present in the learning domain, contributing to ZTab's strong performance on the test datasets. In contrast, supervised models, which are specifically trained on the learning dataset T2D, perform less effectively on the test datasets Limaye and Efthymiou.

%\vspace{-5mm}
\subsubsection{Cross-Ontology Generalization}
\label{sec:cross_ontology}

Table \ref{table:generalization} (right) shows results on Cross-Ontology Generalization where SOTAB\textsubscript{dbp}, SOTAB\textsubscript{sch}, and SOTAB\textsubscript{sch-s} are the learning domains and SOTAB\textsubscript{dbp} is the test domain. 
% Supervised models cannot handle ontology shifts in this setting.
Zero-shot baselines rely solely on the test dataset without any fine-tuning on the learning domain, resulting in the same performance as In-Domain Generalization for SOTAB\textsubscript{dbp}. 

ZTab (LLama3.1-70B,Qwen2.5-7B) significantly outperforms zero-shot models, thanks to leveraging a small number of examples per class to distinguish between similar classes. Additionally, ZTab's adaptability to the test ontology is attributed to (1) incorporating $C_{pred}$ into the prompt to guide the LLM toward correct predictions and (2) applying the label remapping to address syntactic mismatches between predicted classes and classes from the test domain. 
Comparing the ZTab fine-tuned on SOTAB\textsubscript{sch} and SOTAB\textsubscript{sch-s} with the ZTab fine-tuned on the test  SOTAB\textsubscript{dbp}, all being tested on SOTAB\textsubscript{dbp}, the former suffers only a slightly lower performance, showing ZTab’s robustness to ontology shifts.

\begin{table*}[t]
\vspace{-3mm}
\caption{Details of Comparing ZTab-performance with closed-source based models for In-domain generalizations.}
\vspace{-2mm}
\label{table:in_domain_perf}
\centering
\begin{tabular}{@{}lllllllllll@{}}
 &  & 
\multicolumn{7}{c}{\textbf{In-Domain}} \\ 
\cmidrule(lr){3-9} 
\textbf{Category} & \textbf{Model} & 
\textbf{WikiTable} & \textbf{SOTAB\textsubscript{sch}} & 
\textbf{SOTAB\textsubscript{sch-s}} & \textbf{SOTAB\textsubscript{dbp}} & 
\textbf{T2D} & \textbf{Efthymiou} & \textbf{Limaye} \\

\midrule
\multirow{8}{*}{Zero-shot} 
 & Chorus (GPT-3.5) & 15.3 & 60.3 & 60.3 & 64.8 & 83.4 & 82.5 & 87.4 \\
 & GPT-3.5-based & 28.3 & 64.0 & 64.0 & 70.0 & 89.4 & 82.7 & 90.4 \\
%\cmidrule(l){2-9}
 & CENTS (GPT-3.5) & 25.6 & 49.7 & 49.7 & 53.7 & 88.0 & 87.9 & 89.5 \\
 & \textbf{ZTab-per (GPT-3.5,GPT-3.5)} & \textbf{42.3} & \textbf{77.3} & \textbf{77.0} & \textbf{77.4} & \textbf{96.2} & \textbf{92.4} & \textbf{92.5} \\
\cmidrule(l){2-9}
 & CENTS (GPT-4o-mini) & 33.0 & 59.8 & 59.8 & 61.1 & 92.4 & 85.6 & 86.8 \\
 & \textbf{ZTab-per (GPT-4o-mini,GPT-4o-mini)} & \textbf{44.2} & \textbf{77.6} & \textbf{77.1} & \textbf{77.6} & \textbf{96.2} & \textbf{93.4} & \textbf{93.9} \\
 \cmidrule(l){2-9}
 & CENTS (GPT-4o) & 35.7 & 78.1 & 78.1 & 74.5 & 94.7 & 91.3 & 92.1 \\
 & \textbf{ZTab-per (GPT-4o,GPT-4o)} & \textbf{48.8} & \textbf{78.6} & \textbf{78.9} & \textbf{78.9} & \textbf{97.7} & \textbf{95.3} & \textbf{93.9} \\
  \cmidrule(l){2-9}
& CENTS (GPT-4.1-mini) & 38.9 & 76.0 & 76.0 & 73.6 & 96.2 & 85.3 & 80.7 \\
& \textbf{ZTab-per (GPT-4.1-mini,GPT-4.1-mini)} & \textbf{48.9} & \textbf{79.2} & \textbf{78.2} & \textbf{79.3} & \textbf{97.7} & \textbf{94.7} & \textbf{93.9} \\
%& \textbf{ZTab-universal (GPT-4.1-mini)} & \textbf{42.1} & \textbf{77.6} & \textbf{77.6} & \textbf{76.6} & \textbf{97.7} & \textbf{94.7} & \textbf{93.0} \\
\midrule
% \multirow{2}{*}{\textcolor{brown}{\shortstack{Supervised\\(Reference)}}} 
%  & \textcolor{brown}{Doduo (BERT)} & \textcolor{brown}{75.2} & \textcolor{brown}{86.3} & \textcolor{brown}{81.1} & \textcolor{brown}{85.2} & \textcolor{brown}{91.1} & \textcolor{brown}{-} & \textcolor{brown}{-} \\
%  & \textcolor{brown}{ArcheType\textsubscript{FT} (Llama-7B)} & \textcolor{brown}{76.7} & \textcolor{brown}{85.1} & \textcolor{brown}{83.0} & \textcolor{brown}{83.6} & \textcolor{brown}{88.0} & \textcolor{brown}{-} & \textcolor{brown}{-} \\
\multirow{2}{*}{\shortstack{Supervised\\(Reference)}} 
 & Doduo (BERT)& 75.2 & 86.3 & 81.1 & 85.2 & 91.1 & - & - \\
 & ArcheType\textsubscript{FT} (Llama-7B) & 76.7 & 85.1 & 83.0 & 83.6 & 88.0 & - & - \\
\end{tabular}

\vspace{5mm}

\caption{Details of Comparing ZTab-performance with closed-source based models for Cross-domain (left) and cross-ontology (right) generalization.}
\vspace{-2mm}
% Supervised models are not applicable in the cross-ontology setting.}
% Each ZTab variant corresponds to the same backbone model (Llama-7B, Mistral-7B, or Qwen2.5-7B). 
% }
\label{table:generalization_perf}
\centering
%\resizebox{\textwidth}{!}{%
\begin{tabular}{@{}l l  ccc | ccc@{}}
%\toprule
 &  & 
\multicolumn{3}{c}{\textbf{Cross-Domain (Test Datasets)}} & 
\multicolumn{3}{c}{\textbf{Cross-Ontology (Learning Datasets)}} \\
\cmidrule(lr){3-5} \cmidrule(lr){6-8}
\textbf{Category} & \textbf{Model} & 
\textbf{T2D} & \textbf{Efthymiou} & \textbf{Limaye} & 
\textbf{SOTAB\textsubscript{dbp}} & \textbf{SOTAB\textsubscript{sch}} & \textbf{SOTAB\textsubscript{sch-s}} \\
\midrule

\multirow{8}{*}{Zero-shot}
 & Chorus (GPT-3.5) & 83.4 & 82.5 & 87.4 & 64.8 & 64.8 & 64.8 \\
 & GPT-3.5-based & 89.4 & 82.7 & 90.4 & 70.0 & 70.0 & 70.0 \\
 & CENTS (GPT-3.5) & 88.0 & 87.9 & 89.5 & 53.7 & 53.7 & 53.7 \\
 & \textbf{ZTab-per (GPT-3.5,GPT-3.5)} & \textbf{96.2} & \textbf{91.7} & \textbf{92.1} & \textbf{77.4} & \textbf{77.1} & \textbf{76.7} \\
\cmidrule(l){2-8}
 & CENTS (GPT-4o-mini) & 92.4 & 85.6 & 86.8 & 61.1 & 61.1 & 61.1 \\
 & \textbf{ZTab-per (GPT-4o-mini,GPT-4o-mini)} & \textbf{96.2} & \textbf{92.8} & \textbf{93.9} & \textbf{77.6} & \textbf{77.0} & \textbf{77.3} \\
 \cmidrule(l){2-8}
 & CENTS (GPT-4o) & 94.7 & 91.3 & 92.1 & 74.5 & 74.5 & 74.5 \\
 & \textbf{ZTab-per (GPT-4o,GPT-4o)} & \textbf{97.7} & \textbf{94.8} & \textbf{93.9} & \textbf{78.9} & \textbf{78.1} & \textbf{78.5} \\
\cmidrule(l){2-8}
 & CENTS (GPT-4.1-mini) & 96.2 & 85.3 & 80.7 & 73.6 & 73.6 & 73.6 \\
 & \textbf{ZTab-per (GPT-4.1-mini,GPT-4.1-mini)} & \textbf{97.7} & \textbf{94.7} & \textbf{93.9} & \textbf{79.3} & \textbf{78.8} & \textbf{77.7} \\
 %& \textbf{ZTab-universal (GPT-4.1-mini)} & \textbf{97.7} & \textbf{94.7} & \textbf{93.0} & \textbf{76.6} & \textbf{76.6} & \textbf{76.6} \\
\midrule
% \multirow{4}{*}{\textcolor{brown}{\shortstack{Supervised\\(Reference)}}}
%  & \textcolor{brown}{Doduo (BERT)} & \textcolor{brown}{91.1} & \textcolor{brown}{63.2} & \textcolor{brown}{64.5} & \textcolor{brown}{--} & \textcolor{brown}{--} & \textcolor{brown}{--} \\
%  & \textcolor{brown}{ArcheType\textsubscript{FT} (Llama-7B)} & \textcolor{brown}{88.0} & \textcolor{brown}{53.0} & \textcolor{brown}{57.6} & \textcolor{brown}{--} & \textcolor{brown}{--} & \textcolor{brown}{--} \\
%  & \textcolor{brown}{ColNet (Deep learning)} & \textcolor{brown}{94.7} & \textcolor{brown}{61.9} & \textcolor{brown}{59.7} & \textcolor{brown}{--} & \textcolor{brown}{--} & \textcolor{brown}{--} \\
%  & \textcolor{brown}{HNN (Deep learning)} & \textcolor{brown}{96.6} & \textcolor{brown}{65.0} & \textcolor{brown}{74.6} & \textcolor{brown}{--} & \textcolor{brown}{--} & \textcolor{brown}{--} \\
\multirow{4}{*}{\shortstack{Supervised\\(Reference)}}
 & Doduo (BERT)& 91.1 & 63.2 & 64.5 & -- & -- & -- \\
 & ArcheType\textsubscript{FT} (Llama-7B)& 88.0 & 53.0 & 57.6 & -- & -- & -- \\
 & ColNet (Deep learning)& 94.7 & 61.9 & 59.7 & -- & -- & -- \\
 & HNN (Deep learning)& 96.6 & 65.0 & 74.6 & -- & -- & -- \\
%\bottomrule
\end{tabular}%
%}

\vspace{-4mm}
\end{table*}

\subsection{ZTab-performance}
\label{sec:performnace}
This section provides the data-specific comparison for the summary results for ZTab-performance in Table III.  Table \ref{table:in_domain_perf} details the comparison for In-Domain generalization, and 
Table \ref{table:generalization_perf} details the comparison for Cross-Domain and Cross-Ontology Generalizations. In general, ZTab-performance outperforms corresponding baselines across all datasets.

\subsection{Class-specific Analysis}
\begin{figure*}[t]
\centering
\includegraphics[width=1\textwidth]{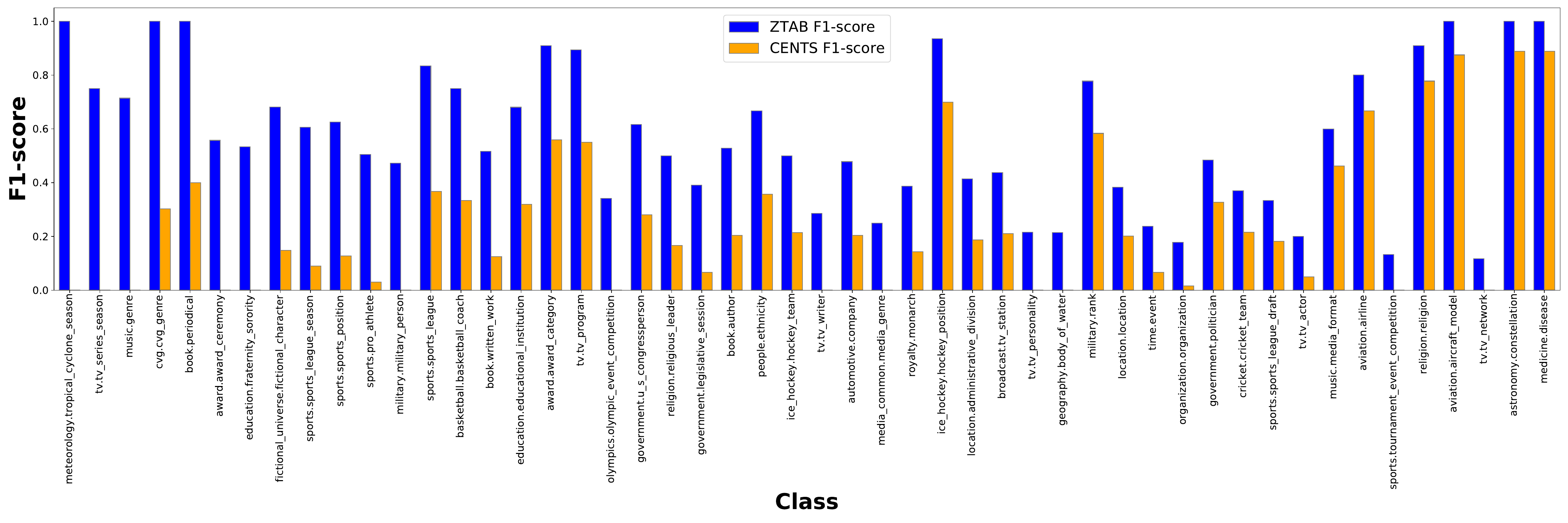} 
\vspace{-5mm}
\caption{Top 50 classes in WikiTable dataset ranked by improvement of ZTab-performance (GPT-4.1-mini,GPT-4.1-mini) over baseline CENTS (GPT-4.1-mini)}
\label{fig: top_50}
\vspace{-3mm}
\end{figure*}

We dive into the performance comparison on individual classes using the WikiTable dataset that has largest and most diverse set of semantic types among all benchmarks. Figure~\ref{fig: top_50} shows the F1-scores for top-50 classes ranked by the improvement of ZTab-performance (GPT-4.1-mini,GPT-4.1-mini) over the strongest baseline, CENTS (GPT-4.1-mini), on this dataset. 

We observe two main groups of classes where ZTab-performance shows substantial improvement. The first group are rare or domain-specific classes such as \textit{meteorology.tropical\_cyclone\_season} and \textit{fictional\_universe.fictional\_character}, for which CENTS often fails with near-zero F1-scores. By fine-tuning annotation LLM with pseudo-tables that include examples of these classes, ZTab exposes model to rare types and improves recognition.  

The second group includes semantically similar classes such as \textit{music.genre} vs. \textit{cvg.cvg\_genre}, and \textit{film.film\_genre} vs. \textit{cvg.cvg\_genre}. ZTab's structured fine-tuning helps the model better distinguish between related but distinct semantic domains by learning subtle contextual differences from generated examples.  For instance, test columns labeled as \textit{music.genre} contain values like \{Techno, Electronica, Rave, Dance, Garage Mix, ...\}, while \textit{cvg.cvg\_genre} columns include \{Action-Adventure, Action RPG, Platformer, Survival Horror, First-person shooter, ...\}, and \textit{film.film\_genre} columns include \{Comedy, Teen, Animated film, Romantic comedy, ...\}. CENTS incorrectly predicts \textit{cvg.cvg\_work} for nearly all \textit{music.genre} columns and for about 40\% of \textit{film.film\_genre} columns, indicating confusion across domains. In contrast, ZTab better distinguishes these categories by leveraging pseudo-tables that capture both intra-class consistency and inter-class distinctions.

\subsection{Ablation Analysis} 
\label{sec:ablation}
In this section, we analyze the effect of different components of ZTab, namely, table generation, pseudo-tables, class prototype size $e$, schema sampling rating $r$, prompt design, row size $k$, and fine-tuning cost. Our analysis focuses on In-Domain Generalization. We also study ontology alignment for Cross-Ontology generalization to evaluate the effect of prompt design and label-remapping when the class list at inference differs from training. We use open-source models (Llama3.1-70B,Qwen2.5-7B) as the prototype LLM and annotation LLM for ZTab-privacy, which offers the flexibility of control. 

\textbf{Table Generation:} Table \ref{tab:table_generation} compares two ways of generating pseudo-tables for fine-tuning of ZTab. 
One uses class prototypes to construct pseudo-tables dynamically, as for ZTab-privacy. The other generates full tables using Llama3.1-70B prior to fine-tuning, as suggested in TabGen \cite{berkovitch2025generating}. 
ZTab clearly performs better on all datasets, showing the better generalization power of the pseudo-tables dynamically constructed using class prototypes.

\begin{table}[h]
\vspace{-3mm}
\centering
\caption{Effect of table generation on micro-F1 score.}
\label{tab:table_generation}
\vspace{-2mm}
\resizebox{\columnwidth}{!}{%
\begin{tabular}{l?cc}
\textbf{Dataset} & ZTab & ZTab with pseudo-tables generated by TabGen \cite{berkovitch2025generating} \\
\thickhline
WikiTable & \textbf{34.1} & 12.9 \\
SOTAB\textsubscript{sch} & \textbf{76.9} & 39.5 \\
SOTAB\textsubscript{sch-s} & \textbf{75.1} & 35.7 \\
SOTAB\textsubscript{dbp} & \textbf{76.2} & 50.3 \\
T2D & \textbf{96.2} & 91.9 \\
\end{tabular}%
}
\vspace{-2mm}
\end{table}

\textbf{Pseudo-table Settings:} 
We examine three settings of
incorporating class prototypes ($P$) and table schemas ($S$) for fine-tuning the annotation model $M_a$: \textbf{ZTab\textsubscript{\textit{w/o $P$}}}: No class prototype $P$ is used in fine-tuning, therefore, the original LLM $M_a$ is applied in Algorithm \ref{alg:prediction} to predict semantic types. This setting is similar to traditional zero-shot learning. \textbf{ZTab\textsubscript{\textit{w/o $S$}}}: Only class prototypes $P$, not the table schemas $S$, are used in fine-tuning. Consequently, for each semantic type a pseudo-table with a single column is created at step 2 of Algorithm \ref{alg:learning}. Under this setting, learning does not leverage the multi-column context as in multi-column pseudo-tables. \textbf{ZTab}: Both class prototypes $P$ and table schema collection $S$ are used, which is the default ZTab in Algorithm \ref{alg:learning}. 

Table \ref{table:data_availability} compares the micro-F1 scores under these settings. 
ZTab\textsubscript{\textit{w/o $P$}} has the poorest performance due to the lack of a learning phase. 
ZTab\textsubscript{\textit{w/o $S$}} has a significant improvement by including the learning phase. The absence of table schema collection $S$ means that the learning cannot leverage multi-column relationships, but it still benefits from single-column pseudo-tables created from class prototypes during the learning process. 
ZTab performs best as it utilizes both class prototypes and table schema collection to take the full advantage of the table's global context provided by all its columns for predicting the type of the target column. Note that, due to the absence of training data in Limaye and Efthymiou, table schema collection $S$ is unavailable, so the performance of ZTab is the same as ZTab\textsubscript{\textit{w/o $S$}}.

\begin{table}[h]
\centering
\vspace{-3mm}
\caption{Effect of pseudo-table settings on micro-F1 score.}
\vspace{-2mm}
\label{table:data_availability}
\begin{tabular}{l?ccc}
Dataset & ZTab & ZTab\textsubscript{\textit{w/o $S$}} & ZTab\textsubscript{\textit{w/o $P$}}   \\
%\hline
\thickhline
WikiTable & \textbf{34.1} & 28.4 &  3.9  \\
SOTAB\textsubscript{sch} & \textbf{76.9} & 71.1 & 18.6     \\
SOTAB\textsubscript{sch-s} & \textbf{75.1} & 71.1 & 18.6    \\
SOTAB\textsubscript{dbp} & \textbf{76.2} & 72.7 & 31.3    \\
T2D & \textbf{96.2} & 93.9 & 67.4    \\
Efthymiou & \textbf{91.3} & 91.3 & 51.5 \\
Limaye & \textbf{92.5} & 92.5 & 76.0   \\
\end{tabular}
%\vspace{-5mm}
\end{table}

\textbf{Class Prototype Size $e$:}
Table \ref{table:description_length} presents the performance of ZTab under varying class prototype sizes $e$ (i.e., 500 (i.e., All), 50, 25, 12, and 6). The best performance is achieved with the full size, as more examples of classes lead to more diverse pseudo-training tables, which improve the model's ability to generalize. However, ZTab demonstrates robust performance even with as few as 6 examples per class, by leveraging the extensive knowledge encoded in LLM's pre-training on large textual corpora. For the best performance, we recommend the full class prototype size $e=500$ for more example diversity. Note that the large prototype size does not necessarily lead to a longer fine-tuning time because the pseudo-table size is determined by the row size $k$, which is 3 in our experiments.

\begin{table}[h]
\vspace{-1mm}
\centering
\caption{Effect of class prototype size $e$ on micro-F1 score. }
\vspace{-2mm}
\label{table:description_length}
\begin{tabular}{l?ccccc}
\multicolumn{1}{c}{} & \multicolumn{4}{c}{\textbf{Class Prototype Size}}\\
%\hline
Dataset & All & 50 & 25 & 12 & 6 \\
\hline
WikiTable &\textbf{34.1} & 32.8& 32.4& 31.2& 30.9 \\
SOTAB\textsubscript{sch} & \textbf{76.9} & 75.9 & 75.4 & 74.7 & 73.6  \\
SOTAB\textsubscript{sch-s} & \textbf{75.1} & 74.2 & 73.3 & 72.5 & 72.8 \\
SOTAB\textsubscript{dbp} & \textbf{76.2} & 74.4 & 74.2 & 73.3 & 72.1 \\
T2D & \textbf{96.2} & 95.9 & 95.4 & 95.6 & 95.2 \\
Efthymiou & \textbf{91.3} & 90.0 & 89.6 & 88.7 & 87.0 \\
Limaye & \textbf{92.5} & 91.9 & 90.8 & 91.5 & 90.9 \\
\end{tabular}

\vspace{-5mm}
\end{table}

\textbf{Prompt Design:}\label{sec:prompt_design}
We explore alternative prompt designs and prediction methods of the \textbf{PromptConstruction} function. The table can be presented in prompt either column-by-column or row-by-row, and prediction method can be either predicting all columns together or predicting one target column at a time.   
Table \ref{table:prompt_design} compares the performance of ZTab with these alternatives. Efthymiou and Limaye datasets are excluded from this analysis since they only have tables with a single column.

The best performance is observed with the column-by-column presentation and target column prediction. The column-by-column presentation allows ZTab to focus on the context of each column individually, which simplifies the learning because the values within each column present examples of the same semantic type. In contrast, the row-by-row presentation introduces values of different semantic types on each row, which makes it harder for the row-based reading to capture the relationships between columns. 
When predicting all columns together, ZTab's performance tends to decrease, particularly when using a smaller annotation LLM $M_a$ like Qwen, because the model may generate an incorrect number of semantic types for a table (e.g., predicting four or six types for a table with five columns). Furthermore, even if ZTab detects the correct semantic types, it may not align them correctly with the corresponding columns.

\begin{table}[h]
\centering
\vspace{-3mm}
\caption{Effect of prompt design on micro-F1 score.}
\vspace{-2mm}
\label{table:prompt_design}
\resizebox{\columnwidth}{!}{
\begin{tabular}{@{}l?cccc@{}}
%\thickhline
Presentation & col-by-col & col-by-col & row-by-row & row-by-row \\
Prediction   & target     & all        & target     & all        \\
\thickhline
WikiTable             & \textbf{34.1} & 31.3 & 32.2 & 28.8 \\
SOTAB\textsubscript{sch}   & \textbf{76.9} & 74.0 & 74.3 & 71.5 \\
SOTAB\textsubscript{sch-s} & \textbf{75.1} & 73.5 & 74.1 & 71.1 \\
SOTAB\textsubscript{dbp}   & \textbf{76.2} & 75.2 & 74.3 & 72.0 \\
T2D                   & \textbf{96.2} & 94.4 & 95.1 & 92.4 \\
%\thickhline
\end{tabular}
}
\vspace{-2mm}
\end{table}

\textbf{Schema Sampling Ratio $r$}: 
Table \ref{table:header_ratio} shows how ZTab performs with different schema sampling ratios $r$. 
We consider WikiTable and three SOTAB datasets that have a large table schema collection $S$. 
The worst performance (SOTAB datasets) occurs at the small sampling ratio of 1\% due to too few schemas used in each epoch. Increasing to 2.5\% improves results, but a further increase provides little additional benefit. This is because many schemas in \( S \) are redundant (see Table \ref{tab:schema-coverage}), sampling all of them is unnecessary, and a 2.5\% sampling ratio captures enough variety of schemas over the specified number of epochs and provides wide and most likely all class coverage. For WikiTable, which contains more schemas and higher redundancy, a smaller 0.5\% ratio is enough.

\begin{table}[h]
\vspace{-3mm}
\caption{Number of total and unique schemas in datasets.}
\vspace{-2mm}
\label{tab:schema-coverage}
\centering
\begin{tabular}{l?rr}
%\toprule
\textbf{Dataset} & \textbf{\#Schemas} & \textbf{\#Unique Schemas} \\
\thickhline
WikiTable & 397,098 & 9,849 \\
SOTAB\textsubscript{sch} & 44,769 & 4,189 \\
SOTAB\textsubscript{sch-s} & 10,631 & 1,643 \\
SOTAB\textsubscript{dbp} & 37,631 & 1,780 \\
T2D & 160 & 64 
%\bottomrule
\end{tabular}

\vspace{5mm}
%\end{table}

%\begin{table}[h]
\centering
\caption{Effect of schema sampling ratios $r$ on micro-F1 score.}
\vspace{-2mm}
% T2D, Limaye, and Efthymiou are excluded since they have no/a few table schemas.}
\label{table:header_ratio}
\begin{tabular}{l?cccccc}
\multicolumn{1}{c}{} & \multicolumn{6}{c}{\textbf{Schema sampling ratio}}\\
Dataset& $10\%$ & $7.5\%$ & $5\%$ & $2.5\%$ & $1\%$ &$0.5\%$ \\
\thickhline
WikiTable & 34.8 & 33.6 & 34.5 & 32.9 & 33.8&34.1 \\
SOTAB\textsubscript{sch}  &  76.7& 77.3 & 76.9 & 76.9 & 74.6 &73.6 \\
SOTAB\textsubscript{sch-s}  & 75.4 & 74.8 & 75.3 & 75.1 & 73.4 &72.8\\
%SOTAB\textsubscript{dbp}  & 0.76 &0.76  & 0.74& 0.76&0.74 & 0.72\\
SOTAB\textsubscript{dbp}  & 76.6 & 75.9 & 76.3 &76.2 & 74.3 & 73.7\\
%T2D &  &  &  &  &  \\
%\hline
\end{tabular}

\end{table}

\textbf{Row Size $k$}: All our experiments for ZTab are based on $k=3$, i.e., all pseudo-tables have 3 rows. Larger \( k \) values add little benefit in performance but increases computational cost.

\begin{figure}[t]
\centering
\includegraphics[trim=0 0 0 1.5cm, clip, width=1\columnwidth, height=4.0cm]{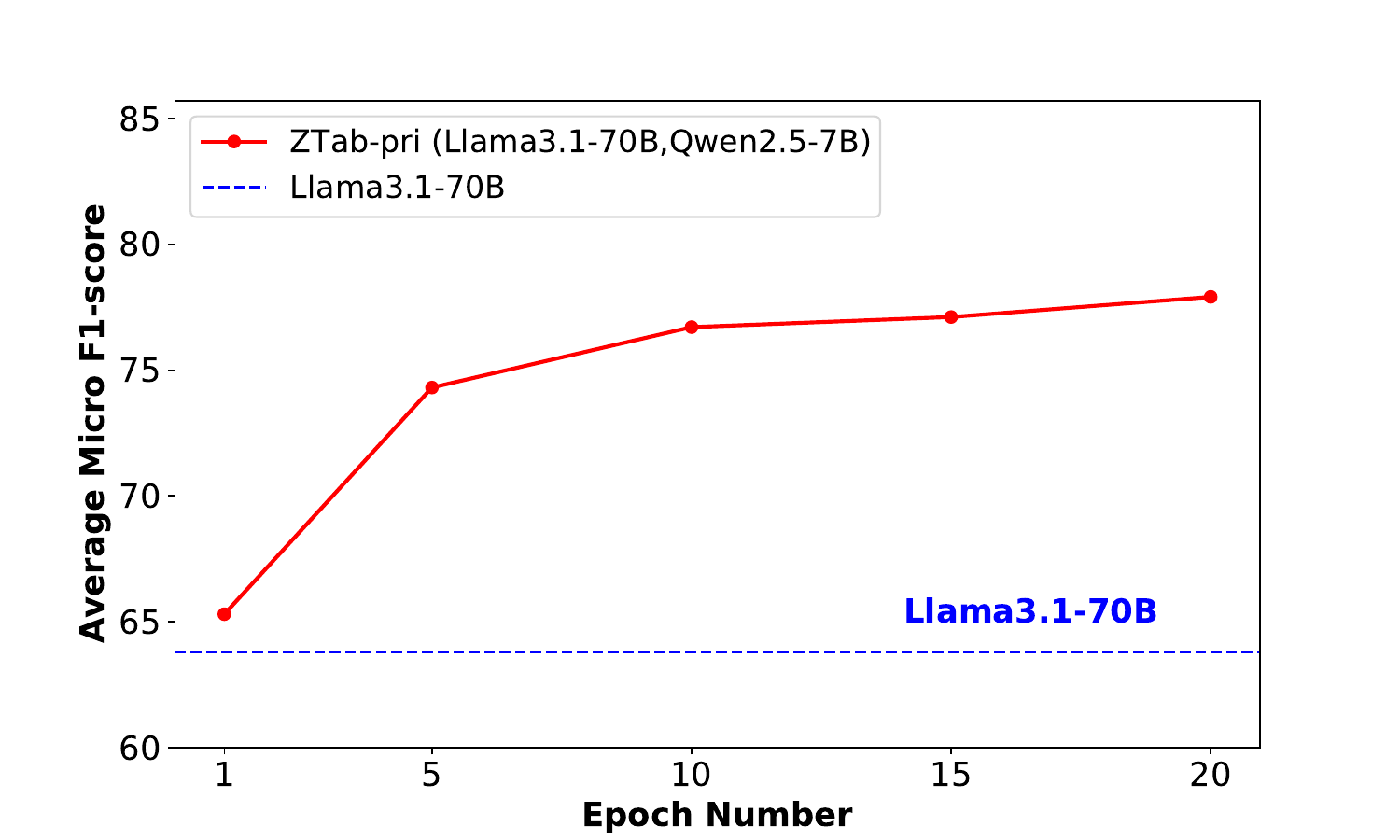} 
\vspace{-7mm}
\caption{Effect of fine-tuning epochs on micro F1 score (avg. over datasets).}
\label{fig: training_time}
\vspace{-6mm}
\end{figure}

\textbf{Fine-tuning Cost}: Compared to zero-shot baselines, ZTab pays in fine-tuning cost, 20 epochs in our experiments: approximately a few minutes for T2D, Limaye, and Efthymiou, and approximately 2, 3, 5, and 7 hours for SOTAB\textsubscript{sch-s}, SOTAB\textsubscript{dbp}, SOTAB\textsubscript{sch}, and WikiTable, respectively. For performance-critical applications, this cost is justified by the performance gain over zero-shot baselines (i.e.,  average improvement of 23.5\%, 1.4\%, 9.5\% at in-domain, cross-domain, and cross-ontology over the strongest baseline). 
Figure \ref{fig: training_time} plots average F1-score (across all datasets) against number of epochs (In-domain Generalization). ZTab outperforms the strongest open-source baseline (Llama3.1-70B) after one epoch and achieves most of its performance gain by the fifth epoch. This suggests that the fine-tuning time of ZTab can be substantially reduced by a brief fine-tuning cycle while providing a substantial improvement on performance. 
In addition, this fine-tuning cost is amortized over time because no retraining is needed 
as long as test domain $C_{pred}$ falls into three generalization scenarios, even when test data distribution or test domain has changed. 

%\textcolor{brown}{
The environmental impact can be determined from fine-tuning runtime and local electricity mix. According to \cite{lannelongue2021green}, for a fixed hardware setup and similar utilization, electricity use is proportional to runtime and CO2 can be estimated as $\text{CO2e} \approx (P_{\text{tot}} \times T)\times I$, where $P_{\text{tot}}$ is the average total power draw during fine-tuning (kW), $T$ is the wall-clock fine-tuning time (hours), and $I$ is the location-based grid carbon intensity (kgCO2e/kWh). Since $P_{\text{tot}}$ and $I$ vary by hardware and location, we report fine-tuning time as a proxy and provide the above formula for deployment-specific CO2 accounting.

%}

%\begin{comment}
\textbf{Ontology Alignment:} For cross-ontology generalization, Algorithm~\ref{alg:prediction} deals with aligning ontology shift (from \( C_{learn} \) to \( C_{pred} \)) at inference time by the prompt construction (line 1) and the label remapping (line 4). To better understand the contribution of these components, we compare two variants of ZTab as follows: \textbf{ZTab}: The default configuration as in Algorithm~\ref{alg:prediction} where prompts are constructed using \( C_{pred} \), as in line 1, and remapping is applied, as in line 4. \textbf{ZTab\textsuperscript{\textit{only-remapping}}}: the prompts are constructed using \( C_{learn} \), i.e., replacing $C_{pred}$ with $C_{learn}$ on line 1, and the remapping on line 4 is used to align predictions with \( C_{pred} \).  Table~\ref{tab:ontology_alignment} shows that ZTab\textsuperscript{\textit{only-remapping}} performs poorly, indicating that relying solely on post-processing mapping from \( C_{learn} \) to \( C_{pred} \) fails to handle differences in ontology definitions. ZTab, combining both prompting and remapping, achieves a high accuracy.

\begin{table}[h]
\vspace{-3mm}
\centering
\small
\caption{Effect of label remapping on micro-F1 under ontology shift.}
\vspace{-2mm}
\label{tab:ontology_alignment}
\begin{tabular}{l?cc}
%\thickhline
Learning Dataset & \textbf{ZTab} & ZTab\textsuperscript{\textit{only-remapping}} \\
\thickhline
SOTAB\textsubscript{sch} & \textbf{75.2} & 55.6 \\
SOTAB\textsubscript{sch-s} & \textbf{74.4}  & 54.9 \\
\end{tabular}
\end{table}
\vspace{-5mm}
%\end{comment}

\section{Conclusion}
We presented ZTab, a novel \textit{domain-based zero-shot} framework for column type annotation to overcome the limitations of existing zero-shot and supervised models. ZTab eliminates the need for user-labeled training data while learning tabular structures and distinguishing between similar classes. This is achieved by incorporating domain information at schema level (i.e., class list and table schemas) to generate pseudo-tables for fine-tuning a pre-trained LLM. Pseudo-tables are constructed using LLM-generated class prototypes, independently of data distribution, allowing fine-tuned model to be zero-shot because no retraining is needed when test data or test domain has changed for three generalization scenarios (i.e., In-Domain, Cross-Domain, and Cross-Ontology). Our domain-based zero-shot ZTab provides a trade-off between performance and zero-shot through domain configuration.

\section{Acknowledgment}
The work of Ke Wang is supported in part by a discovery grant from Natural Sciences and Engineering Research Council of Canada.

\section{AI-generated content acknowledgment}
LLMs have been used to lightly edit the entire manuscript.

\bibliographystyle{IEEEtran}
\bibliography{sample_file}

\end{document}